\documentclass[10pt,twocolumn,letterpaper]{article}

\usepackage{cvpr}
\usepackage{times}
\usepackage{epsfig}

\usepackage{graphicx}
\usepackage{caption}
\usepackage{subcaption}
\usepackage{url}
\usepackage{calc}

\usepackage{amsmath}
\usepackage{amsfonts}
\usepackage{amssymb}
\usepackage{amsthm}

\usepackage{algorithmic}
\usepackage{algorithm}

\usepackage{color}
\usepackage{tikz}
\usepackage{wrapfig}
\usepackage{ulem}
\usepackage{enumitem}
\usepackage{dsfont}

\newtheorem{theorem}{Theorem}%[section]
\newtheorem{corollary}{Corollary}%[section]
\DeclareMathOperator{\vMF}{\text{vMF}}

\DeclareMathOperator{\Dir}{\text{Dir}}

\DeclareMathOperator{\Cat}{\text{Cat}}

\DeclareMathOperator*{\argmax}{arg\,max}

\newcommand{\grad}[1]{\ensuremath{\nabla_{#1}}}
\newcommand{\largeproptau}{\ensuremath{\tau^{(*)}}}
\newcommand{\proptau}[1]{\ensuremath{\overset{\largeproptau}{#1}}}

\newcommand{\dt}[1]{\ensuremath{{{\Delta t}_{#1}}}}

\newcommand{\Bessel}[1]{\ensuremath{\text{I}_{#1}}}

\newcommand{\bx}{\textbf{x}}
\newcommand{\bmu}{{\boldsymbol\mu}}

\newcommand{\bz}{\textbf{z}}

\newcommand{\SD}{\mathbb{S}^{D-1}}

\newcommand{\indices}{\mathcal{I}}

\usepackage[pagebackref=false, breaklinks=true, letterpaper=true, colorlinks, bookmarks=false]{hyperref}
\newcommand{\kmeans}{\emph{k}-means}
\newcommand{\spkm}{spkm}
\newcommand{\DPvMFmeans}{DP-vMF-means}

\newcommand{\DONE}[1]{}

\cvprfinalcopy % *** Uncomment this line for the final submission

\def\cvprPaperID{440} % *** Enter the CVPR Paper ID here

\ifcvprfinal\fi

\makeatletter
\renewcommand{\paragraph}{%
  \@startsection{paragraph}{4}%
  {\z@}{1ex \@plus 1ex \@minus .2ex}{-1em}%
  {\normalfont\normalsize\bfseries}%
}
\makeatother

\begin{document}

%%%%%%%%% TITLE
\title{Small-Variance Nonparametric Clustering on the Hypersphere}

\author{Julian Straub $\quad$ Trevor Campbell $\quad$ Jonathan
  P.~How $\quad$ John W.~Fisher III\\
CSAIL and LIDS, Massachusetts Institute of Technology\\
{\tt\small \{jstraub, fisher\}@csail.mit.edu, \{tdjc, jhow\}@.mit.edu}
}

\maketitle

\begin{abstract}
  Structural regularities in man-made environments reflect in the
distribution of their surface normals.  Describing these surface normal
distributions is important in many computer vision applications, such
as scene understanding, plane segmentation, and regularization of 3D
reconstructions.  Based on the small-variance limit of Bayesian
nonparametric von-Mises-Fisher (vMF) mixture distributions, we propose two
new flexible and efficient $k$-means-like clustering algorithms for directional data
such as surface normals. The first, DP-vMF-means, is a batch clustering
algorithm derived from the Dirichlet process (DP) vMF mixture.  Recognizing
the sequential nature of data collection in many applications, we
extend this algorithm to DDP-vMF-means, which infers temporally
evolving cluster structure from streaming data. Both algorithms
naturally respect the geometry of directional data, which lies on the
unit sphere.  We demonstrate their performance on synthetic directional
data and real 3D surface normals from RGB-D sensors.  While our
experiments focus on 3D data, both algorithms generalize to
high dimensional directional data such as protein backbone
configurations and semantic word vectors.

%Structural regularities in man-made environments reflect in the distribution of their surface normals. Describing these surface normal distributions is important in many computer vision applications, such as scene understanding, plane segmentation, and regularization of 3D reconstructions. Based on the small-variance limit of Bayesian nonparametric von-Mises-Fisher (vMF) mixture distributions, we propose two new flexible and efficient k-means-like clustering algorithms for directional data such as surface normals. The first, DP-vMF-means, is a batch clustering algorithm derived from the Dirichlet process (DP) vMF mixture. Recognizing the sequential nature of data collection in many applications, we extend this algorithm to DDP-vMF-means, which infers temporally evolving cluster structure from streaming data. Both algorithms naturally respect the geometry of directional data, which lies on the unit sphere. We demonstrate their performance on synthetic directional data and real 3D surface normals from RGB-D sensors. While our experiments focus on 3D data, both algorithms generalize to high dimensional directional data such as protein backbone configurations and semantic word vectors.

\end{abstract}
\section{Introduction}\label{sec:intro}
Man-made environments and objects exhibit clear structural regularities
such as planar or rounded surfaces. These properties are evident on all
scales from small objects such as books, to medium-sized scenes like tables, rooms and
buildings and even to the organization of whole cities. Such regularities can
be captured in the statistics of surface normals that describe the
local differential structure of a shape.  These statistics contain
valuable information that can be used for scene understanding, plane
segmentation, or to regularize a 3D reconstruction.

Inference algorithms in fields such as robotics or augmented reality,
which would benefit from the use of surface normal statistics, are not
generally provided a single batch of data a priori. Instead, they are
often provided a stream of data batches from 
depth cameras. Thus, capturing the surface normal
statistics of man-made structures often necessitates the temporal
integration of observations from a vast data stream of varying cluster
mixtures.  Additionally, such applications pose hard constraints on the
amount of computational power available, as well as tight timing
constraints. 

\begin{figure}
  \centering
  \begin{tikzpicture}
    \node (t0) at (-3,0) {\includegraphics[width=0.2\columnwidth]{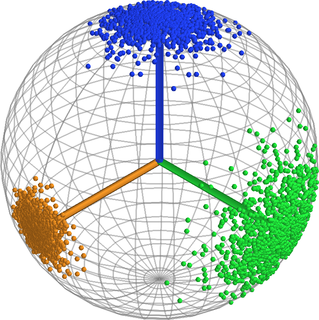}};
    \node at (-3.7,0.8) {\footnotesize$t_0$};
    \node (t1) at (-1,0) {\includegraphics[width=0.2\columnwidth]{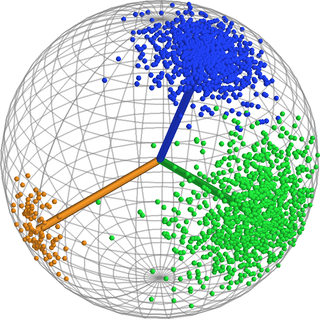}};
    \node at (-1.7,.8) {\footnotesize$t_1$};
    \node at (-2,-1.2) {\footnotesize movement};
    \draw[->,line width=1pt,line join=round,line cap=round] (-2.6,-.86) to[out=330,in=210] (-1.4,-.86);
    \node (t2) at (1,0) {\includegraphics[width=0.2\columnwidth]{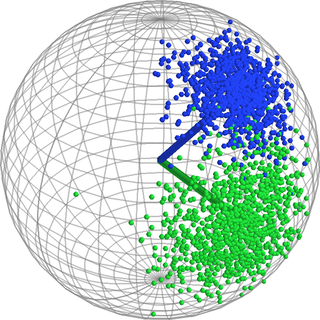}};
    \node at (.3,.8) {\footnotesize$t_2$};
    \node at (0,-1.2) {\footnotesize death};
    \draw[->,line width=1pt,line join=round,line cap=round] (-.6,-.86) to[out=330,in=210] (.6,-.86);
    \node (t3) at (3,0) {\includegraphics[width=0.2\columnwidth]{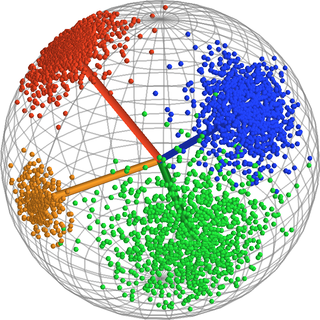}};
    \node at (2.3,.8) {\footnotesize$t_3$};
    \node at (2,-.8) {{\footnotesize birth}};
    \node at (2,-1.2) {{\footnotesize revival}};
    \draw[->,line width=1pt,line join=round,line cap=round] (1.4,-.86) to[out=330,in=210] (2.6,-.86);
  \end{tikzpicture}
  \caption{Evolution of the data distribution on the sphere which we model
  using a dependent Dirichlet process von-Mises-Fisher mixture model
  (DDP-vMF-MM).\label{fig:intro}}
\end{figure}
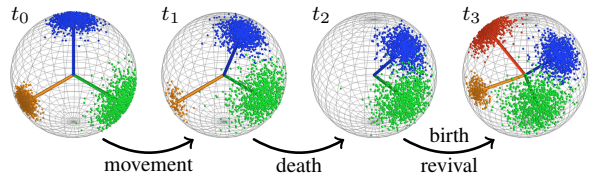

We address these challenges by focusing on flexible Bayesian
nonparametric (BNP) Dirichlet process mixture models (DP-MM) which
describe the distribution of surface normals in their natural space,
the unit sphere in 3D. Taking the small-variance asymptotic limit of
this DP-MM of von-Mises-Fisher (vMF) distributions, we obtain a fast
\emph{k}-means-like algorithm, which we call DP-vMF-means, to perform
nonparametric clustering of data on the unit hypersphere.  Furthermore,
we propose a novel dependent DP mixture of vMF distributions to achieve
integration of directional data into a temporally consistent streaming
model (shown in Fig.~\ref{fig:intro}). Small-variance asymptotic
analysis yields the \emph{k}-means-like DDP-vMF-means algorithm.
Finally, we propose a method, inspired by optimistic concurrency
control, for parallelizing the inherently sequential labeling process
of BNP-derived algorithms.  This allows real-time processing of batches
of ~300k data-points at 30~Hz.

Beyond the aforementioned vision applications, directional data is
ubiquitous in many other fields, including protein backbone
configurations in computational biology, semantic word vectors in
natural language processing, and rotations expressed as quaternions in
robotics. Further, many of those sources are observed as a stream. The
proposed algorithms directly generalize to these other data sources as
well as higher-dimensional spherical data.

\section{Related Work}\label{sec:related}
\paragraph{Directional distributions:} A variety of
distributions~\cite{mardia2009directional} has been proposed in the
field of directional statistics to model data on the unit sphere.
Examples are the antipodal symmetric Bingham
distribution~\cite{bingham1974antipodally}, the anisotropic Kent
distribution~\cite{kent1982fisher}, and the isotropic von-Mises-Fisher
(vMF) distribution~\cite{fisher1995circularData}.  Because of its
comparative simplicity, the vMF distribution is most commonly used.

\paragraph{vMF mixture models:} vMF mixture models (vMF-MM) are
especially popular for modeling and inference purposes.
Banerjee \etal~\cite{banerjee2005clustering} perform Expectation
Maximization (EM) for a finite vMF mixture model to cluster text and
genomic data. 
This method is related to the spherical \kmeans{} (\spkm)
algorithm~\cite{dhillon2001concept}, which can be obtained from a
finite vMF-MM by taking the infinite limit of the concentration
parameter~\cite{banerjee2005clustering}.
Zhong~\cite{zhong2005efficient} extends the spherical \kmeans{}
algorithm to an online clustering framework by performing stochastic
gradient descent on the spkm objective. This approach requires the
number of clusters to be known, does not allow the creation or
deletion of clusters, and heuristically weights the contribution of old
data.
Gopal~\etal~\cite{gopal2014mises} derive variational as well as
collapsed Gibbs sampling inference for finite vMF-MMs, for a finite
hierarchical vMF-MM and for a finite temporally evolving vMF-MM.
The vMF distribution has also been used in BNP and hierarchical MMs.
Bangert \etal~\cite{bangert2010using} formulate an infinite vMF-MM
using a Dirichlet process (DP) prior, but their sampling-based
inference is known to have convergence issues and to be
inefficient~\cite{Jain00asplit-merge}.
Reisinger~\etal~\cite{reisinger2010spherical} formulate a finite,
latent Dirichlet allocation (LDA) model~\cite{blei2003lda} for
directional data using the vMF distribution. 
To the best of our knowledge there are no other related
\emph{k}-means-like algorithms for directional batch data besides the
\spkm{} algorithm. 
In many applications the spherical manifold of the data is ignored and
clustering is performed in the ambient Euclidean space~\cite{Holz11RoboCup}. 

\paragraph{Surface normal modeling:} 
In the context of object and 3D shape representation, extended Gaussian
images (EGI) have been
studied~\cite{horn1984extended,ikeuchi1981recognition}. The EGI of a
surface is the distribution of surface normals where each surface
normal is weighted by the area of the surface it represents.  For
computations the EGI was approximated via a tessellation of the sphere.
A crucial property of EGI is that the representation is invariant to
translation.
Exploiting this property, Makadia~\etal~\cite{makadia2006fully} extract
maxima in the EGI and use the spherical FFT to compute an initial
rotation estimate for point-cloud registration. Note that these maxima
correspond to the cluster centers which the proposed (D)DP-vMF-means
algorithm extracts.
Furukawa~\etal~\cite{furukawa2009manhattan} use a tessellation of the
unit sphere to extract the dominant directions in a scene as part of a
depth regularization algorithm based on the Manhattan World (MW)
assumption~\cite{coughlan1999manhattan}.
Tribel~\etal~\cite{triebel2005using} employ EM to
perform plane segmentation using surface normals in combination with 3D
locations.  Assuming a finite vMF-MM and using a hierarchical
clustering approach, Hasnat~\etal~\cite{hasnat2013hierarchical} model
surface normal distributions.
Straub~\etal~\cite{straub2014mmf} propose a probabilistic model to
describe mixtures of Manhattan frames (MMF) as orthogonally coupled
clusters of surface normals on the sphere. 
In a similar approach,
but without MMF constraints,
Straub~\etal~\cite{straub2015dptgmm} introduce efficient inference for
a DP-MM of clusters in distinct tangent spaces to the sphere.

\paragraph{Small-variance asymptotics:} A principled way to reduce the
computational cost of inference, while still 
drawing on the capabilities of BNP models,
is to apply small-variance analysis to the model.  This technique was first applied
to the DP Gaussian mixture~\cite{jordan2012dpmeans}, resulting in a
\emph{k}-means-like algorithm with a linear regularization on the
number of clusters.
Since then, the technique has been extended to develop asymptotic
algorithms for mixtures with general exponential family
likelihoods~\cite{Jiang12_NIPS}, HMMs with an unknown number of
states~\cite{Roychowdhury13_NIPS}, and dynamic, time-dependent
mixtures~\cite{campbell2013dynamic}.  However, small-variance analysis
has, to date, been limited to discrete and Euclidean spaces.

\section{Von-Mises-Fisher Mixture Models}
The von-Mises-Fisher (vMF)~\cite{fisher1995circularData} distribution
with mean direction $\mu$ and concentration
$\tau$ is an isotropic distribution over the $D$-dimensional
unit hypersphere, $\SD = \{x \in \mathbb{R}^D : x^Tx = 1\}$,
(Fig.~\ref{fig:transitionGeometry}).
Its density is defined as~\cite{bangert2010using}
\begin{align}
\begin{aligned}
  \vMF(x;\mu,\tau) &= Z(\tau)\exp(\tau \mu^Tx)\\
  Z(\tau) &=\tau^{D/2 -1} {(2\pi)}^{- D/2} {\Bessel{D/2-1}(\tau)}^{-1}\,,
\end{aligned}
\end{align} 
where $\mu, x \in \SD$, $\tau>0$ and \Bessel{\nu} is the modified Bessel function~\cite{Abramowitz65} of the first
kind of order $\nu$. 
The conjugate prior for $\mu$ given a fixed $\tau$ is a vMF distribution $\vMF(\mu; \mu_0, \tau_0)$
and the corresponding posterior given data $\bx = \{x_i\}_{i=1}^N$ is
\begin{align} \label{eq:vMFposterior}
\begin{aligned}
  p(\mu| \bx; \tau, \mu_0, \tau_0) &\propto p(\mu; \mu_0, \tau_0)  \prod_{i=1}^N p(x_i|\mu;\tau) \\
  \therefore p(\mu| \bx; \tau, \mu_0, \tau_0)& =  \vMF(\mu;
  \tfrac{\!\!\mu'}{\|\mu'\|_2}, \|\mu'\|_2) \,,
\end{aligned}
\end{align}
where $\mu' = \tau_0 \mu_0 + \tau\sum^N_{i=1} x_i$.

A finite mixture of $K$ $\vMF$ distributions with known concentration $\tau$ 
may be obtained by placing a Dirichlet distribution prior $\Dir(\alpha)$ 
on the mixture weights $\pi$, and a $\vMF$ prior on the mean directions
$\mu_k$. Data points $\bx$ are assigned to clusters via
latent indicator variables $\bz = \{z_i\}_{i=1}^N$.
\begin{align}
\begin{aligned}
  \pi &\sim \Dir(\alpha), \, \, \mu_k  \sim \vMF(\mu_0,\tau_0) \,  \forall k \in \{1,\dots,K\} \\
  z_i &\sim \Cat(\pi), \, \, x_i \sim \vMF(\mu_{z_i},\tau) \,  \forall i \in \{1,\dots,N\} \,.
\end{aligned}
\end{align}
Parallel to the connection between \emph{k}-means and the Gaussian
mixture model in the small-variance asymptotic
limit~\cite{jordan2012dpmeans}, taking $\tau\to\infty$ yields
deterministic updates as also previously noted
in~\cite{banerjee2005clustering}. For completeness of the presentation
we give a detailed derivation in the supplement.

\section{Dirichlet Process vMF-MM}\label{sec:dpvmf}
The Dirichlet process (DP)~\cite{ferguson1973bayesian,Teh10_EML}
has been widely used as a prior for mixture models with a countably-infinite set of
  clusters~\cite{antoniak1974mixtures,bangert2010using,chang13dpmm,neal2000markov}.
  Assuming a base distribution  $\vMF(\mu; \mu_0, \tau_0)$, the DP is
an appropriate prior for a $\vMF$ mixture with an unknown number of
components and known $\vMF$ concentration $\tau$.
Gibbs sampling inference only differs from the finite
Dirichlet vMF-MM in the label sampling step; the mixing weights $\pi$
are integrated out, resulting in the Chinese Restaurant Process
(CRP)~\cite{blackwell1973ferguson,neal2000markov}
\begin{align}
  \begin{aligned}
    &p(z_i =k | \bz_{-i}, \bmu, \bx; \tau, \mu_0, \tau_0) \\
    &\propto\left\{\begin{array}{ll}
      |\indices_k| \vMF(x_i | \mu_k ; \tau) & k \leq K \\
    \alpha \, p(x_i ; \tau, \mu_0, \tau_0) & k=K+1  \,,
  \end{array}\right.
  \end{aligned}\label{eq:dplabelsample}
\end{align}
where $\indices_k$ is the set of data indices assigned to cluster $k$.
Note that the DP concentration parameter, $\alpha > 0$, influences
the likelihood of adding a new clusters. The conjugate prior for
$\mu_k$ yields $p(x_i ; \tau, \mu_0, \tau_0)$ via marginalization:
\begin{align}
    p(x_i ; \tau, \mu_0, \tau_0)
    &= \int \vMF(x_i | \mu_k; \tau) \vMF(\mu_k; \mu_0, \tau_0)
    \:\textrm{d} \mu_k \nonumber\\
    &= \frac{Z(\tau) Z(\tau_0)}{Z(||\tau x_i  + \tau_0 \mu_0||_2 )} \,. \label{eq:priormarginal}
\end{align} 
\subsection{DP-vMF-means\label{sec:dpvmf}}
In this section, we provide a small-variance asymptotic analysis of the
label and parameter update steps of the Gibbs sampling algorithm for
the DP vMF mixture, yielding deterministic updates. As
with \emph{k}-means, the label assignments are computed sequentially
for all datapoints before the means are updated, and the process is
iterated until convergence. Pseudocode can be found in the supplement.

\paragraph{Label Update:} To derive a hyperspherical analog to
DP-means~\cite{jordan2012dpmeans}, consider the limit of the label
sampling step (\ref{eq:dplabelsample}) as $\tau \rightarrow \infty$.
The normalizer $Z(||\tau x_i  + \tau_0 \mu_0||_2)$ approaches
\begin{align}
  \begin{aligned}
  Z(||\tau x_i  + \tau_0 \mu_0||_2) 
  &\proptau{\to} 
  \exp(-\tau)  \,,
\end{aligned}\label{eq:vMFnormalizerLimit}
\end{align}
where overscript $\largeproptau$ denotes proportionality up to a finite
power of $\tau$, and where we have used the fact that as
$\tau\to\infty$, the modified Bessel function of the first kind
satisfies~\cite{Abramowitz65}
\begin{align}
  \Bessel{D/2-1}(\tau) &= \frac{\exp(\tau)}{\sqrt{2\pi \tau}} \left(1 -
  O\left(\frac{1}{\tau}\right) \right)
  \proptau{\to} \exp\left(\tau\right)\,.\label{eq:mbessapprox} 
\end{align} 

To achieve a nontrivial result, the asymptotic behavior of $Z(\tau)$
must be matched by $\alpha$, so let $\alpha = \exp(\lambda \tau)$ to
obtain
\begin{align} \label{eq:marginalAssymptotics}
  \frac{\alpha Z(\tau) Z(\tau_0)}{Z(||\tau x_i  + \tau_0 \mu_0||_2 )} 
  &\proptau{=}  Z(\tau) \exp(\tau(\lambda+1))\, .
\end{align} 
Therefore, as $\tau\to\infty$, the label sampling step becomes
\begin{align}                                                                   
  \!\!\!\!\begin{aligned}   
    &\lim_{\tau \rightarrow \infty} p(z_i=k  | \bz_{-i}, \bmu, \bx; \tau, \mu_0,
    \tau_0) \\
     &= \!\!\lim_{\tau \rightarrow \infty} 
     \!\left\{\begin{array}{ll}
     \!\!\begin{matrix}
       \frac{ |\indices_k| e^{\tau (x_i^T \mu_k - \lambda-1)}}{ \sum_{j=1}^K
       |\indices_j| e^{\tau (x_i^T \mu_j - \lambda -1)} +  c(\tau)} \\
     \end{matrix} & k\leq K \\
  \!\!\begin{matrix}
       \frac{ c(\tau)}{ \sum_{j=1}^K |\indices_j| e^{\tau (x_i^T \mu_j -
       \lambda -1)} +  c(\tau)} \\
     \end{matrix}  & k=K+1 \,,
   \end{array}\right.
  \end{aligned}   
\end{align} 
where we have used that the normalizers $Z(\tau)$ of $\vMF(x_i|
\mu_k;\tau)$ and Eq.~\eqref{eq:marginalAssymptotics} cancel and
$c(\tau) \proptau{=} 1$.  Thus, as $\tau\to\infty$, sampling from
$p(z_i  | \bz_{-i}, \bmu, \bx; \tau, \mu_0, \tau_0)$ is equivalent to the following
assignment rule:
\begin{align}
  \label{eq:dpvMFassignment}
  z_i = \argmax_{k\in\{1, \dots, K+1\}} \left\{\begin{array}{ll}
                                      x_i^T\mu_k & k\leq K\\
                                      \lambda+1 & k=K+1 \,.
                                    \end{array}\right. 
\end{align}
Since $-1 \leq x_i^T\mu_k \leq 1$, the parameter $\lambda$
can be restricted to the set $\lambda \in \left[-2, 0\right]$ without
loss of generality. 
Intuitively $\lambda$ defines the maximum angular spread
$\phi_\lambda$ of clusters about their mean direction, via $\lambda =
\cos(\phi_{\lambda})-1$.
Note, that upon assigning a datapoint to a new cluster,
i.e.~$z_i = K+1$, the mean of that cluster is initialized to
$\mu_{K+1} = x_i$.
Finally, if an observation $x_i$ is the last one in its
cluster, the cluster is removed prior to finding the new label for
$x_i$ using Eq.~(\ref{eq:dpvMFassignment}).  

\paragraph{Parameter Update:} Taking $\tau\to\infty$ in
the parameter posterior for cluster $k$ from
Eq.~\eqref{eq:vMFposterior} causes $\tau_0$ and $\mu_0$
to become negligible. Hence the parameter update becomes:
\begin{align}
  \mu_k = \frac{\sum_{i\in\indices_k}x_i}{\|\sum_{i\in\indices_k}x_i\|_2}  \quad\forall k \in \{1, \dots, K\} \,.
\end{align} 

\paragraph{Objective Function:}
From Eq.~\eqref{eq:dpvMFassignment} we can see that assigning a
datapoint $x_i$ to cluster $k$ provides a score of $x_i^T\mu_k$,
whereas adding a new cluster provides a score of $\lambda+1 -
x_i^T\mu_{K+1} = \lambda$, since new mean directions are initialized
directly to $\mu_{K+1} = x_i$. Hence, the objective function that DP-vMF-means maximizes is
\begin{align}
  J_{\text{DP-vMF}} = \sum_{k=1}^K \sum_{i\in\indices_k} x_i^T\mu_k +
  \lambda K \,.
\end{align}

\section{Dependent Dirichlet Process vMF-MM}\label{sec:ddpvmf}
Suppose now that, in addition to an unknown number $K$ of components,
the $\vMF$ mixture undergoes temporal evolution in discrete timesteps
$t \in \mathbb{N}$ (Fig.~\ref{fig:intro}):
mixture components can move, be destroyed, and new ones can be created
at each timestep. For such a scenario, the dependent Dirichlet process
(DDP)~\cite{maceachern1999dependent,lin2010construction,campbell2013dynamic}
is an appropriate prior over the mixture components and weights. Using
intermediate auxiliary DPs $F_0$ and $F_1$, the DDP constructs a Markov
chain of DPs $G_t$, where $G_{t+1}$ is sampled from $G_t$ as follows:
\begin{enumerate}[noitemsep]
  \item \textbf{(Death)} For each atom $\theta$ in $G_t$, sample from $\text{Bernoulli}(q)$. If the result
    is 1, add $\theta$ to $F_0$. 
  \item \textbf{(Motion)} Replace each $\theta$ in $F_0$ with $\theta' \sim
    T(\theta' | \theta)$.
  \item \textbf{(Birth)} Sample a DP $F_1\sim \text{DP}(\alpha,
    H)$. Let $G_{t+1}$ be a random convex combination of $F_0$ and $F_1$.
\end{enumerate}
There are four parameters in this model: $\alpha > 0$ and $H\!\left(\cdot\right)$, the concentration
parameter and base measure of the innovation process; $q \in (0, 1)$, the
Bernoulli cluster survival probability; and finally $T\!\left(\cdot | \cdot\right)$, the random
walk transition distribution. In the present work, both the base and
random transition distributions are von-Mises-Fisher:
$H\!\left(\mu\right) = \vMF\left(\mu; \mu_0, \tau_0\right)$, and
$T\!\left(\mu | \nu \right) = \vMF\left(\mu; \nu, \xi\right)$.

Suppose at timestep $t$, a new batch of data $\bx$ is observed. Then Gibbs
sampling posterior inference for the DDP mixture, as in the previous sections,
iteratively samples labels and parameters.
Let the set of tracked mean directions from previous timesteps be
$\{\mu_{k0}\}_{k=1}^K$, where $\mu_{k0} \sim \vMF(m_k, \tau_k)$ and $\dt{k}$
denotes the number of timesteps since cluster $k$ was last instantiated 
(i.e.~when it last had data assigned to it).\footnote{Note that all quantities
($\bx$, $\mu_{k0}$, $m_k$, $\tau_k$, $c_k$, $n_k$, etc.) are now time-varying. This
dependence is not shown in the notation for brevity, and all
quantities are assumed to be shown for the current timestep $t$.} 
Then the label sampling distribution is
\begin{align}
  \begin{aligned}
    &\hspace{-.2cm}p(z_i = k | \bmu, \bz_{-i}, \bx) \\
    &\hspace{-.4cm}\propto\!\! \left\{\!\!\begin{array}{ll}
    \alpha\frac{1-q^t}{1-q}p(x_{i}; \mu_0, \tau_0)
                                    & k=K+1\\
                                    (c_k+|\indices_k|)\vMF(x_i| \mu_k; \tau) & k\leq K,
                                    |\indices_k|>0\\
                                    q^{\dt{k}}c_k p(x_i;m_k, \tau_k) & k\leq K, 
                                    |\indices_k|=0 \,,
                              \end{array}\right.
                            \end{aligned} \label{eq:ddpAssignmentDistri}
\end{align}
where $c_{k}$ is the number of observations
assigned to cluster $k$ in past timesteps. 
The parameter sampling distribution is
\begin{align}
  \begin{aligned}
    \hspace{-.25cm}p(\mu_k | \bmu_{-k}, \bz, \bx)\hspace{-.05cm}\propto\hspace{-.05cm} \left\{\hspace{-.2cm}\begin{array}{ll}
        \vMF(\mu_k;\tfrac{\!\!\mu'_k}{\|\mu'_k\|_2}, {\scriptstyle
        \|\mu'_k\|_2}) & \hspace{-.2cm}c_k = 0\\
      p(\mu_k| \bx, \bz; m_k, \tau_k) & \hspace{-.2cm}c_k > 0 \,,
      \end{array}\right.
  \end{aligned}\label{eq:ddpParameterDistri}
\end{align}
where $\mu'_k = \tau_0 \mu_0 + \tau\sum_{i\in \indices_k} x_i$, and
$p(\mu_k| \bx, \bz; m_k, \tau_k)$ is the distribution over the current
cluster $k$ mean direction $\mu_k$ given the assigned data and
the old mean direction $\mu_{k0}$.

\subsection{DDP-vMF-means}
In the following, we analyze the small-variance asymptotics of the DDP-vMF
mixture model. We first derive the label assignment rules, followed by 
the parameter updates.

\paragraph{Label Update:} First, let $\alpha = \exp(\lambda \tau)$, $q =
\exp(Q\tau)$, $\xi = \exp(\beta\tau)$, and $\tau_k = \tau w_k$, with
$\lambda\in\left[-2, 0\right]$ as before, $Q \leq 0$, and $\beta, w_k \geq 0$.
Note that $\lim_{\tau\to\infty}\frac{1-q^t}{1-q}= 1$, and thus the asymptotics
of the label assignment probability for current and new clusters is the same as
in Section \ref{sec:dpvmf}.

Hence, we focus on the assignment of a datapoint to a previously
observed, but currently not instantiated, cluster $k$. During the
$\dt{k}$ timesteps since cluster $k$ was last observed, the mean
direction $\mu_k$ underwent a random $\vMF$ walk $\mu_{k0}\to
\mu_{k1}\to \dots\to \mu_{k\dt{k}}= \mu_k$ with initial distribution
$\mu_{k0} \sim \vMF(\mu_{k0}; m_k, \tau_k)$. Therefore, the
intermediate mean directions $\{\mu_{kn}\}_{n=1}^{\dt{k}}$ must be
marginalized out when computing $p(x_i; m_k, \tau_k)$:
\begin{align}
\begin{aligned}
  &p(x_i ; m_k,\tau_k) \\
  &\begin{array}{l}
    = \idotsint\limits_{\mu_{k0}, \dots, \mu_{k\dt{k}}} 
    \begin{matrix}
      p(x_i |\mu_{k\dt{k}};\tau)\cdot p(\mu_0;m_k,\tau_k) \\
      \cdot \prod_{n=1}^\dt{k} p(\mu_{kn} |\mu_{k(n-1)};\xi)
    \end{matrix} \\
    = Z(\tau) Z(\beta\tau)^\dt{k} Z(w_k\tau) \idotsint\limits_{\mu_{k0} ,\dots,
    \mu_{k\dt{k}}}\exp\left(\tau f\right)
  \end{array}\\
&f\! =\! x_i^T \mu_{k\dt{k}}\! + \beta \sum_{n=1}^\dt{k} \mu_{kn}^T
  \mu_{k(n-1)}\! + w_k
  \mu_{k0}^Tm_k \,.
\end{aligned}\label{eq:transint}
\end{align}

The integration in (\ref{eq:transint}) cannot be computed in closed form;
however, the value of the integral is only of interest in the limit
as $\tau \to \infty$. Therefore, Theorem \ref{thm:laplace}, an extension
of Laplace's approximation to general differentiable manifolds,
may be used to obtain an exact formula.
\begin{theorem}[Manifold Laplace Approximation]
  Suppose $M\subset \mathbb{R}^n$ is a bounded $m$-dimensional differentiable manifold 
  and $f\!:\!\mathbb{R}^n\!\!\to\!\mathbb{R}$ is a smooth function on $M$. Further,
  suppose $f$ has a unique global maximum on $M$, $x^\star = \argmax_{x\in M}
  f(x)$. Then
  \begin{align}
    \lim_{\tau\to\infty} \frac{\int_M e^{\tau f(x)}}
    {\left(\frac{2\pi}{\tau}\right)^m\left|\det
    U^T\grad{}^2f(x^\star)U\right|^{\frac{1}{2}}
e^{\tau f(x^\star)}} = 1 \,,
  \end{align}
  where $U\in\mathbb{R}^{n\times m}$ is a matrix whose columns
  are an orthonormal basis for the tangent space of $M$ at $x^\star$.\label{thm:laplace}
\end{theorem}

\begin{proof}
See the supplementary material. The general technique of this proof is to
transform coordinates between the manifold and its tangent plane
  using the exponential map~\cite{Docarmo1992}, 
  and then apply the multidimensional Laplace approximation in the
  transformed Euclidean space.
\end{proof}

\begin{corollary}\label{cor:laplace}
  Given a smooth function $f\!:\!\left(\mathbb{R}^D\right)^N \!\!\to\! \mathbb{R}$, with
  a unique global maximum over the $N$-product of $\nobreak{(D-1)}$-spheres 
  $x^\star \in \left(\mathbb{S}^{D-1}\right)^N$, 
  \begin{align}
    \int_{\left(\mathbb{S}^{D-1}\right)^N} e^{\tau
    f(x)} \proptau{\to} e^{\tau f(x^\star)}\,.
  \end{align}
\end{corollary}

\begin{proof}
  This is Theorem \ref{thm:laplace} applied to
  $\left(\mathbb{S}^{D-1}\right)^N$. 
\end{proof}

Using Corollary \ref{cor:laplace} and the limiting approximation of the
modified Bessel function (\ref{eq:mbessapprox}) in equation (\ref{eq:transint}) yields 
the following asymptotic behavior as $\tau \to \infty$:
\begin{align}
  \begin{aligned}
  p(x_i ; m_k, \tau_k) &\proptau{\to} 
  \exp\left(\tau(f^\star -1 - \beta\dt{k} -w_k) \right) \,.
\end{aligned}\label{eq:LaplacevMFTransition}
\end{align} 

\begin{figure}
  \centering
   \begin{subfigure}{0.27\columnwidth}
     \includegraphics[width=\textwidth]{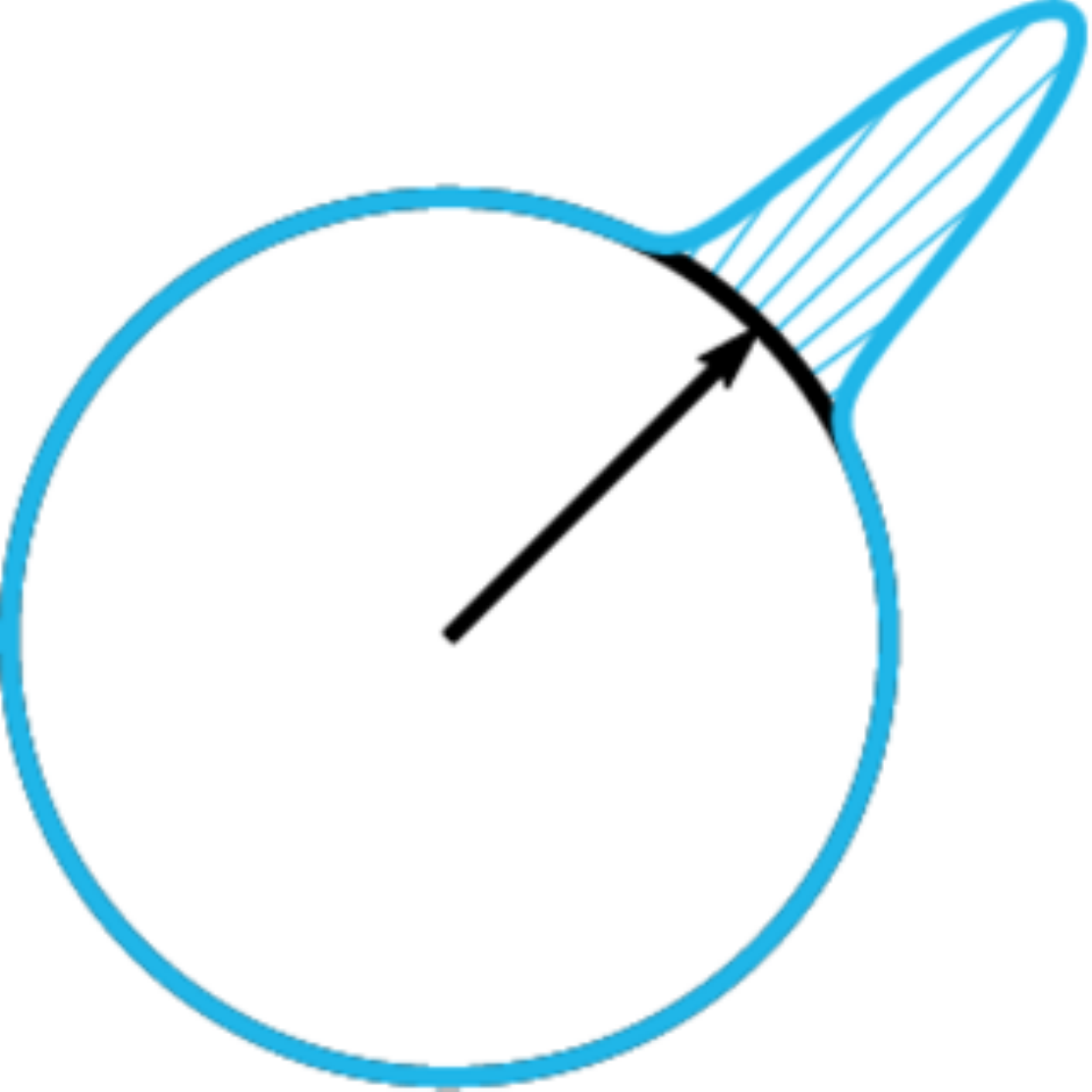}
    \caption{$\tau=100$}
  \end{subfigure}
   \begin{subfigure}{0.27\columnwidth}
     \includegraphics[width=\textwidth]{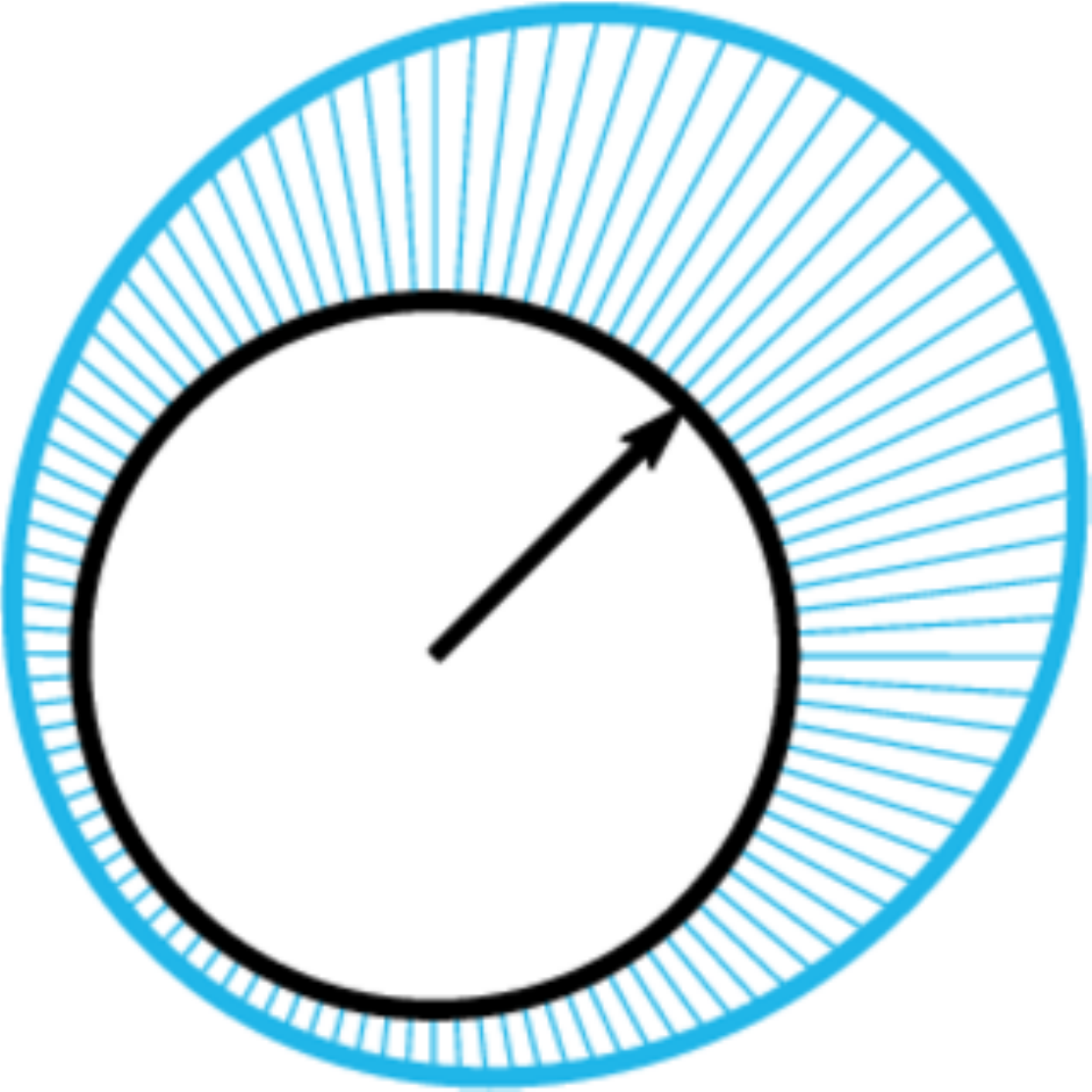}
    \caption{$\tau=1$}
  \end{subfigure}
  \hspace{0.1cm}
  \begin{subfigure}{0.40\columnwidth}
    \scalebox{0.8}{
    \begin{tikzpicture}
  % points on circle
  \draw[anchor=west] (50:2) node {$m_k$};
  \draw[anchor=west] (30:2) node {$\mu_{k0}$};
  \draw[anchor=west] (13:2) node {$\cdot$};
  \draw[anchor=west] (7.5:2) node {$\cdot$};
  \draw[anchor=west] (2:2) node {$\cdot$};
  \draw[anchor=west] (345:2) node {$\mu_{k\dt{k}}$};
  \draw[anchor=west] (325:2) node {$\tfrac{\bar{x}_k}{||\bar{x}_k||_2}$};
  % segments and angles
  \draw[-,black,line width = 1] (0:0) to (50:2);
  \draw  (40:1.2) node {$\theta$};
  \draw[-,black,line width = 1] (0:0) to (30:2);
  \draw  (22.5:1.4) node {$\phi$};
  \draw[-,black,line width = 1] (0:0) to (15:2);
  \draw[] (10:1.4) node {$\cdot$};
  \draw[] (7.5:1.4) node {$\cdot$};
  \draw (5:1.4) node {$\cdot$};
  \draw[-,black,line width = 1] (0:0) to (0:2);
  \draw  (352.5:1.4) node {$\phi$};
  \draw[-,black,line width = 1] (0:0) to (345:2);
  \draw  (335:1.2) node {$\eta$};
  \draw[-,black,line width = 1] (0:0) to (325:2);

  \draw[anchor=south]  (60:.2) node {$\zeta$};
  % circle 
  \draw[dashed,black,line width = 1] (0:2) arc [radius=2,start angle=0, end angle=60];
  \draw[dashed,black,line width = 1] (320:2) arc [radius=2,start angle=320, end angle=360];
  % angle for theta
  \draw[black,line width = 1] (30:1.4) arc [radius=1.4,start angle=30, end angle=50];
  % angles for phi s
  \draw[black,line width = 1] (0:1) arc [radius=1,start angle=0, end angle=30];
  \draw[black,line width = 1] (0:1) arc [radius=1,start angle=0, end angle=30];
  \draw[black,line width = 1] (345:1) arc [radius=1,start angle=345, end angle=360];
  % angle for eta  
  \draw[black,line width = 1] (325:1.4) arc [radius=1.4,start angle=325, end angle=345];
  % angle for zeta
  \draw[black,line width = 1] (325:.6) arc [radius=.6,start angle=325, end angle=360];
  \draw[black,line width = 1] (0:.6) arc [radius=.6,start angle=0, end angle=50];
\end{tikzpicture}
  }
  \end{subfigure}
  \caption{Left: 2D vMF distributions. Right: geometry of the maximum likelihood
    setting  of $\mu_{k0}, \mu_{k1}, \dots,
\mu_{k\dt{k}}$ for the transition distribution. 
\label{fig:transitionGeometry}}
\end{figure}
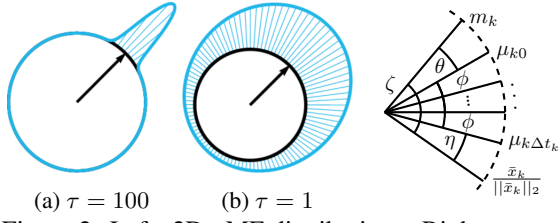

The only remaining unknown in the asymptotic expression, $f^\star$, can be found via constrained optimization 
\begin{align}
\begin{aligned}
  \hspace{-.3cm}\max_{\{\mu_{kn}\}_{n=1}^{\dt{k}}} \; &  x_i^T \mu_{k\dt{k}} + \beta
  \sum_{n=1}^\dt{k} \mu_{kn}^T \mu_{k(n-1)} + w_k \mu_{k0}^T m_k\; \\
  & \text{ s.t. } \mu_{kn}^T\mu_{kn} = 1 \; \forall n \in \{ 0,\dots, \dt{k} \} \,.
\end{aligned}\label{eq:expopt}
\end{align}
The optimization (\ref{eq:expopt}) has a closed-form solution:
\begin{align}
  \begin{aligned}
    \mu_{k0} &= \frac{w_k m_k + \beta \mu_{k1}}{|| w_k m_k + \beta \mu_{k1}||_2}\\
\mu_{kn} &= \frac{\mu_{k(n+1)} + \mu_{k(n-1)}}{||\mu_{k(n+1)} +
\mu_{k(n-1)}||_2} \; { \scriptstyle \forall n \in \{1,\dots, \dt{k}-1 \} }\\
\mu_{k\dt{k}} &= \frac{x_i + \beta \mu_{k(\dt{k}-1)}}{||x_i + \beta
  \mu_{k(\dt{k}-1)}||_2}\,.
  \end{aligned}
\end{align}
These ternary relationships enforce that the optimal vMF
mean directions along the random walk lie on the geodesic between $m_k$ and $x_i$. 
Therefore, this walk can be described geometrically by three angles, as shown in
Fig.~\ref{fig:transitionGeometry} (with $\bar{x}_k = x_i$): the angle $\phi$ between consecutive $\mu_{kn}$, the angle $\eta$ between $x_i$
and $\mu_{k \dt{k}}$, and the angle $\theta$ between $m_k$ and $\mu_{k0}$.
Given these definitions, standard trigonometry yields a set of three equations
in $\phi$, $\eta,$ and $\theta$:
\begin{align}
  \begin{aligned}
      w_k \sin(\theta^\star) = \beta \sin(\phi^\star) = \sin(\eta^\star) \\ 
      \zeta = \theta^\star + \dt{k}  \phi^\star + \eta^\star =
      \arccos(m_k^T x_i ) \,,
    \end{aligned}\label{eq:etathetaphi}
\end{align} 
where $\zeta$ is the full angle between $x_i$ and $m_k$.
Since (\ref{eq:etathetaphi}) cannot be solved in closed-form, Newton's method
is used  to compute $\phi^\star$, $\theta^\star$, and  $\eta^\star$, which in turn
determines $f^\star$:
\begin{align}
  f^\star = w_k \cos(\theta^\star) + \beta \dt{k} \cos(\phi^\star) +
  \cos(\eta^\star) \,.
\end{align}

Returning to (\ref{eq:LaplacevMFTransition}), the transition asymptotics  are
\begin{align}
  p(x_i ; m_k;\tau_k) \proptau{\to}\exp\left(\hspace{-.2cm}\begin{array}{c}
    \tau w_k(\cos(\theta^\star)-1)\\
  +\tau \beta\dt{k}(\cos(\phi^\star)-1)\\
  +\tau (\cos(\eta^\star)-1)\end{array}\hspace{-.2cm}\right)\,.
\end{align} 
Substituting this into Eq.~\eqref{eq:ddpAssignmentDistri} with the earlier
definition $q=\exp(\tau Q)$, and taking the limit $\tau\to\infty$
yields the assignment rule $z_i = \argmax_{k} J_k$, where
\begin{align}
  \hspace{-.2cm}J_k\! =\! \left\{\hspace{-.2cm}\begin{array}{l l}
    \lambda +1  &  k = K+1\\  
    \mu_k^T x_i  &  k\leq K, |\indices_k| > 0\\
      \hspace{-.1cm}\left(\begin{aligned}  \dt{k} \beta &( \cos (\phi^\star ) -1) \\
        + w_k&( \cos(\theta^\star)-1)  \\ 
  + \cos&(\eta^\star) + \Delta t_{k} Q \end{aligned} \right)
      &k\leq K, |\indices_k| =  0 \,.
      \end{array}\right. \label{eq:labelupdate}
\end{align}

Note that if $Q\dt{k} < \lambda$, cluster $k$ can be removed
permanently as it will never be revived again.
Furthermore, for any $Q \leq \lambda$ all clusters are removed after each
timestep, and the algorithm reduces to DP-vMF-means.

\paragraph{Parameter Update:}
The parameter update rule for DDP-vMF-means comes from the asymptotic
behavior of (\ref{eq:ddpParameterDistri}) as $\tau \to \infty$. The
analysis for any new cluster is the same as that in Section
\ref{sec:dpvmf}, so our focus is again on the transitioned mean
direction posterior $p(\mu_{k\dt{k}}| \bx, \bz; m_k, \tau_k)$ (recall
that $\mu_k=\mu_{k\dt{k}}$ in the definition of the random $\vMF$
walk). This distribution can be expanded, similarly to
(\ref{eq:transint}), as:
\begin{align}
  \hspace{-.2cm}\begin{aligned}
    &p(\mu_{k\dt{k}}| \bx, \bz; m_k, \tau_k) \\
  &\begin{array}{l}
    = \idotsint\limits_{\mu_{k0}, \dots, \mu_{k(\dt{k}-1)}} 
    \begin{matrix}
      p(\bx |\mu_{k\dt{k}};\tau)\cdot p(\mu_0;m_k,\tau_k) \\
      \cdot \prod_{n=1}^{\dt{k}} p(\mu_{kn} |\mu_{k(n-1)};\xi)
    \end{matrix} \\
    = Z(\tau)^{|\indices_k|} Z(\beta\tau)^\dt{k} Z(w_k\tau) \hspace{-.3cm}\idotsint\limits_{\mu_{k0} ,\dots,
    \mu_{k(\dt{k}-1)}}\hspace{-.2cm}\exp\left(\tau f\right)
  \end{array}\\
  &f\! =\! \sum_{i\in\indices_k}x_i^T \mu_{k\dt{k}}\! + \beta \sum_{n=1}^\dt{k} \mu_{kn}^T
  \mu_{k(n-1)}\! + w_k
  \mu_{k0}^Tm_k \,.
\end{aligned}\label{eq:ddpPosterior}
\end{align}
Define $\bar{x}_k = \sum_{i\in\indices_k}x_i$. 
Once again, applying Corollary \ref{cor:laplace}, the limit $\tau\to\infty$
removes the integrals over the marginalized mean directions. However, in
contrast to the label assignment update, $\mu_{k\dt{k}}$ is not marginalized
out. Therefore, an additional maximization with respect to $\mu_{k\dt{k}}$ to find 
the concentration point of the posterior yields
\begin{align}
  \hspace{-.15cm}\begin{aligned}
  &p(\mu_{k\dt{k}}| \bx, \bz; m_k, \tau_k) \proptau{\to}\\
  &\quad\exp(\tau(f^\star - |\indices_k|
  -\dt{k} \beta - w_k)) \\
  & f^\star = w_k \cos(\theta^\star) + \beta \dt{k} \cos(\phi^\star) + ||\bar{x}_k||_2 \cos(\eta^\star)\,.
\end{aligned}\label{eq:paramupdateasym}
\end{align}
Analyzing the geometry of the geodesic between $\bar{x}_k/\|\bar{x}_k\|_2$ and $m_k$ (Fig.~\ref{fig:transitionGeometry}) there exist $\phi^\star$, $\theta^\star$ and
$\eta^\star$ such that
  \begin{align}
    \begin{aligned}
      w_k \sin(\theta^\star) = \beta \sin(\phi^\star) = ||\bar{x}_k||_2
      \sin(\eta^\star) \\ 
      \zeta = \theta^\star + \dt{k}  \phi^\star + \eta^\star = \arccos(m_k^T
      \tfrac{\bar{x}_k}{\|\bar{x}_k\|_2}) \,,
    \end{aligned}\label{eq:paramupdate}
  \end{align}
  which can be solved via Newton's method. Given the solution, 
  $\mu_{k}$ can be obtained by rotating
  $\tfrac{\bar{x}_k}{\|\bar{x}_k\|_2}$ by angle $\eta^\star$ on the geodesic
  shown in Fig.~\ref{fig:transitionGeometry} towards
  $m_k$,
\begin{align}
  \hspace{-.5cm}\begin{aligned}
    \mu_{k} &= R(\eta^\star) \tfrac{\bar{x}_k}{ ||\bar{x}_k||_2}\,.
  \end{aligned} 
\end{align} 

\paragraph{Weight Update:} 
After the iteration of label and parameter updates has converged, the
weight $w_k$ must be updated for all clusters to reflect
the new uncertainty in the mean direction of cluster $k$. 
This can be done by examining (\ref{eq:paramupdateasym}): Since
at the maximum of a $\vMF(\mu; m_k, w_k\tau)$ density, $\exp(\tau w_k m_k^T\mu)
= \exp(\tau w_k)$, $w_k$ is updated to $f^\star$.

\begin{figure*}
  \centering
  \begin{subfigure}[c]{0.32\textwidth}
  \centering
    \includegraphics[width=\textwidth]{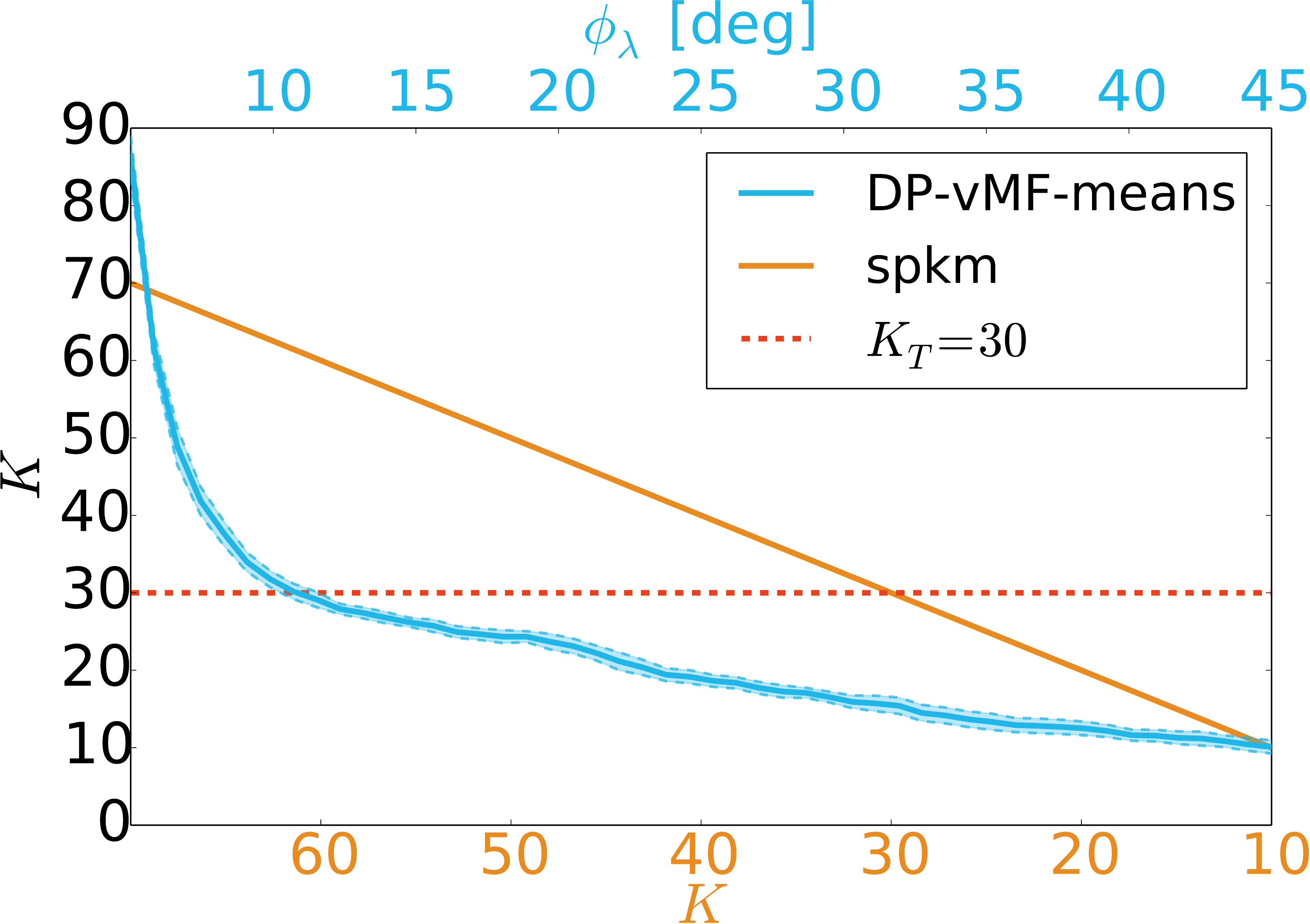}
  \end{subfigure}
  \begin{subfigure}[c]{0.32\textwidth}
  \centering
    \includegraphics[width=\textwidth]{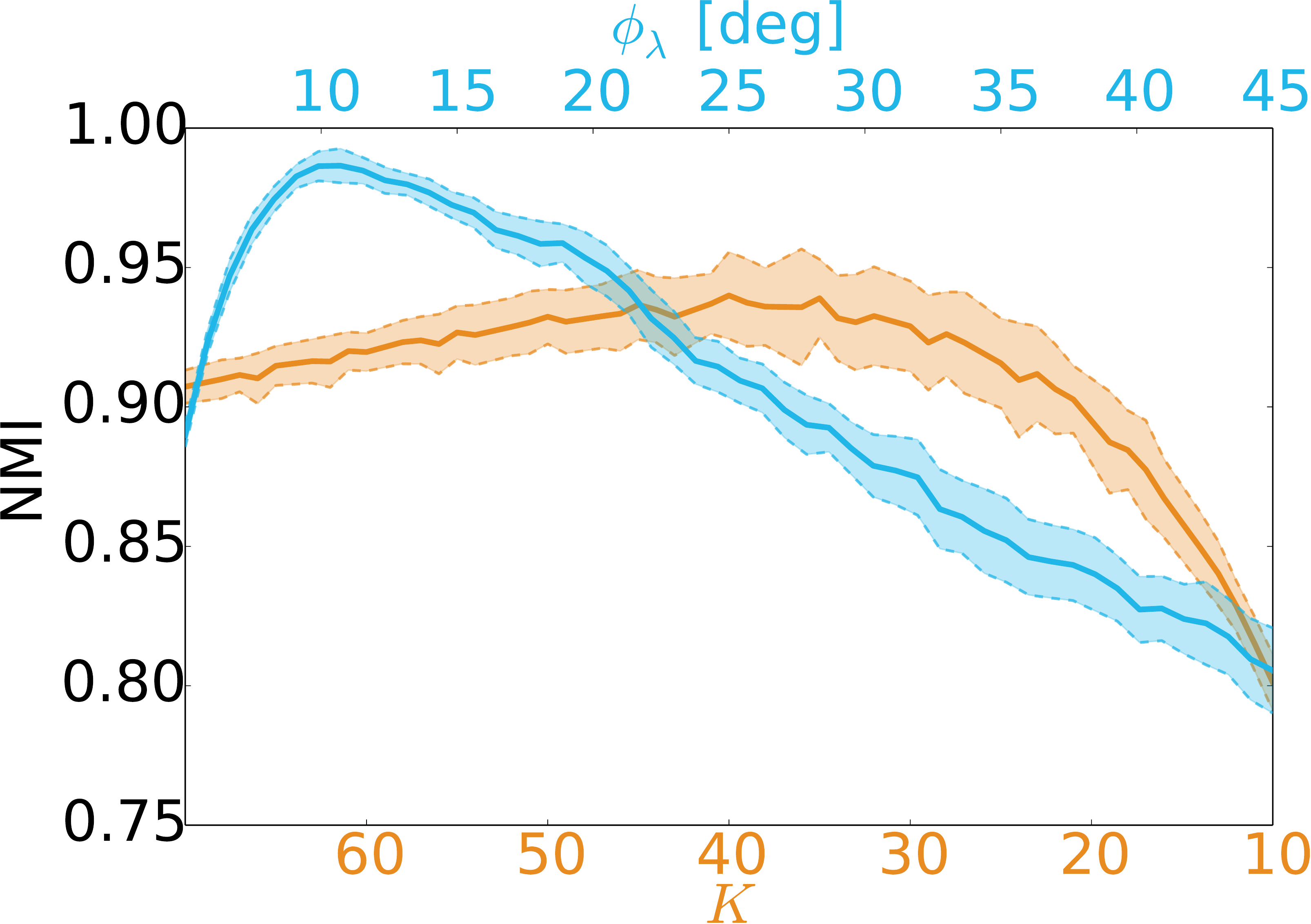}
  \end{subfigure}
  \begin{subfigure}[c]{0.32\textwidth}
  \centering
    \includegraphics[width=\textwidth]{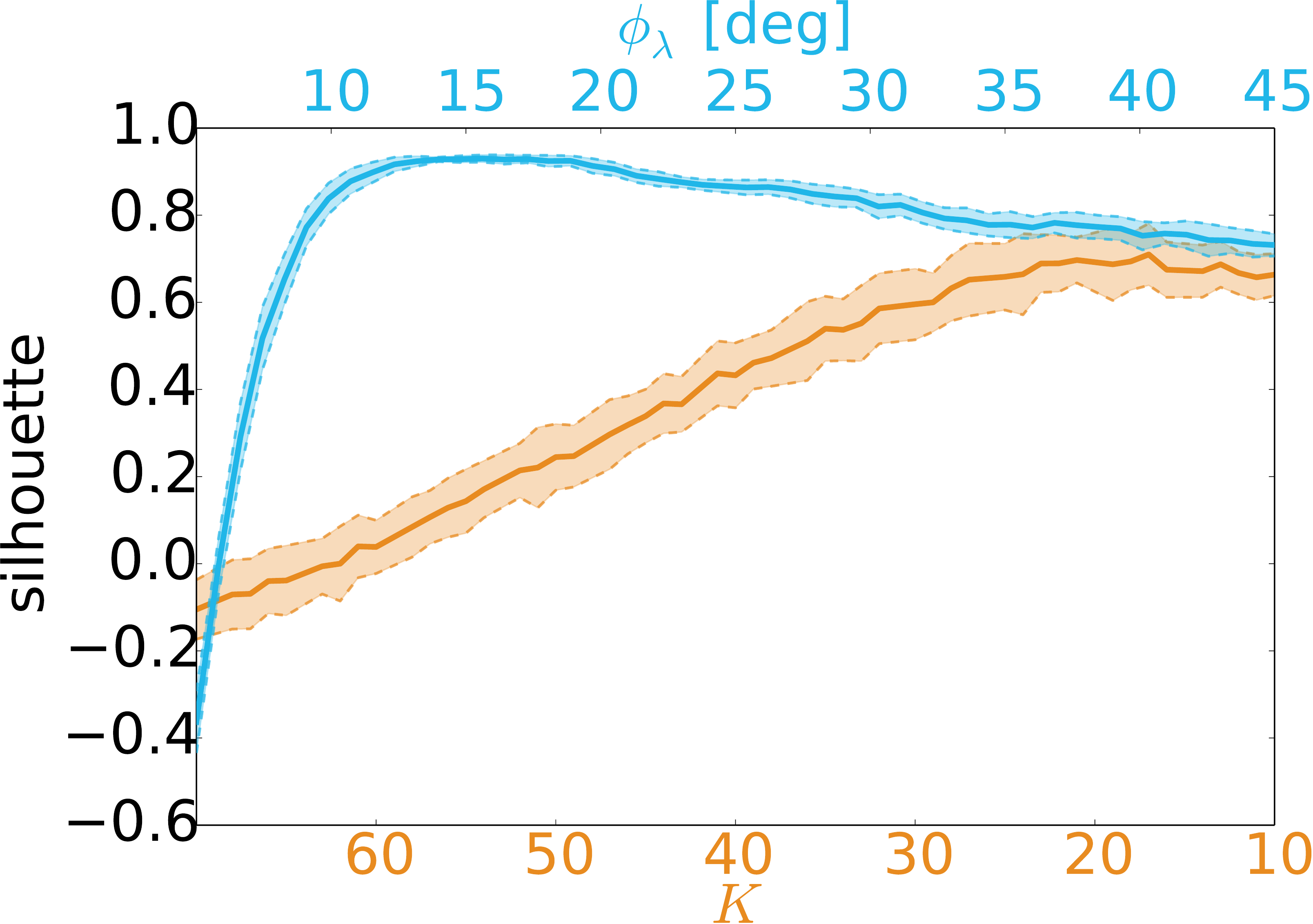}
  \end{subfigure}
  \caption{Comparison of the spkm and the \DPvMFmeans{} clustering algorithms on synthetic spherical data with
    $K_\text{T} =30$ clusters.  Note \DPvMFmeans{}' higher maximum normalized
    mutual information (NMI) as well as silhouette score. 
    \label{fig:synthEval}}
\end{figure*}
\section{Optimistic Iterated Restarts (OIR)}
In our implementation of the algorithm we pay special attention to
speed and parallel execution to enable real-time performance for
streaming RGB-D data.

Observe that the main bottleneck of 
DP-based hard clustering algorithms,
such as the proposed (D)DP-vMF-means, DP-means~\cite{jordan2012dpmeans} or
Dynamic means~\cite{campbell2013dynamic},
is the inherently sequential assignment of labels:
due to the creation of new clusters, the label assignments depend on
all previous assignments.  While this is a key feature of the streaming
clustering algorithms, it poses a computational hindrance.
We address this issue with an optimistic parallel label assignment
procedure inspired by techniques for database concurrency
control~\cite{pan2013optimistic}.

First, we compute assignments in parallel (e.g.~on a GPU).  If all
datapoints were assigned only to instantiated clusters, we output the
labeling. Otherwise, we find the lowest observation id $i$ that
modified the number of clusters, apply the modification, and
recompute the assignments for all observations $i' > i$ in parallel.
Thus, per data-batch, DP-vMF-means restarts once per new cluster, while
DDP-vMF-means restarts once for each new or revived cluster.

\begin{figure}
  \begin{subfigure}[c]{0.16\textwidth}
    \vspace{0.12cm}
    \includegraphics[width=\textwidth]{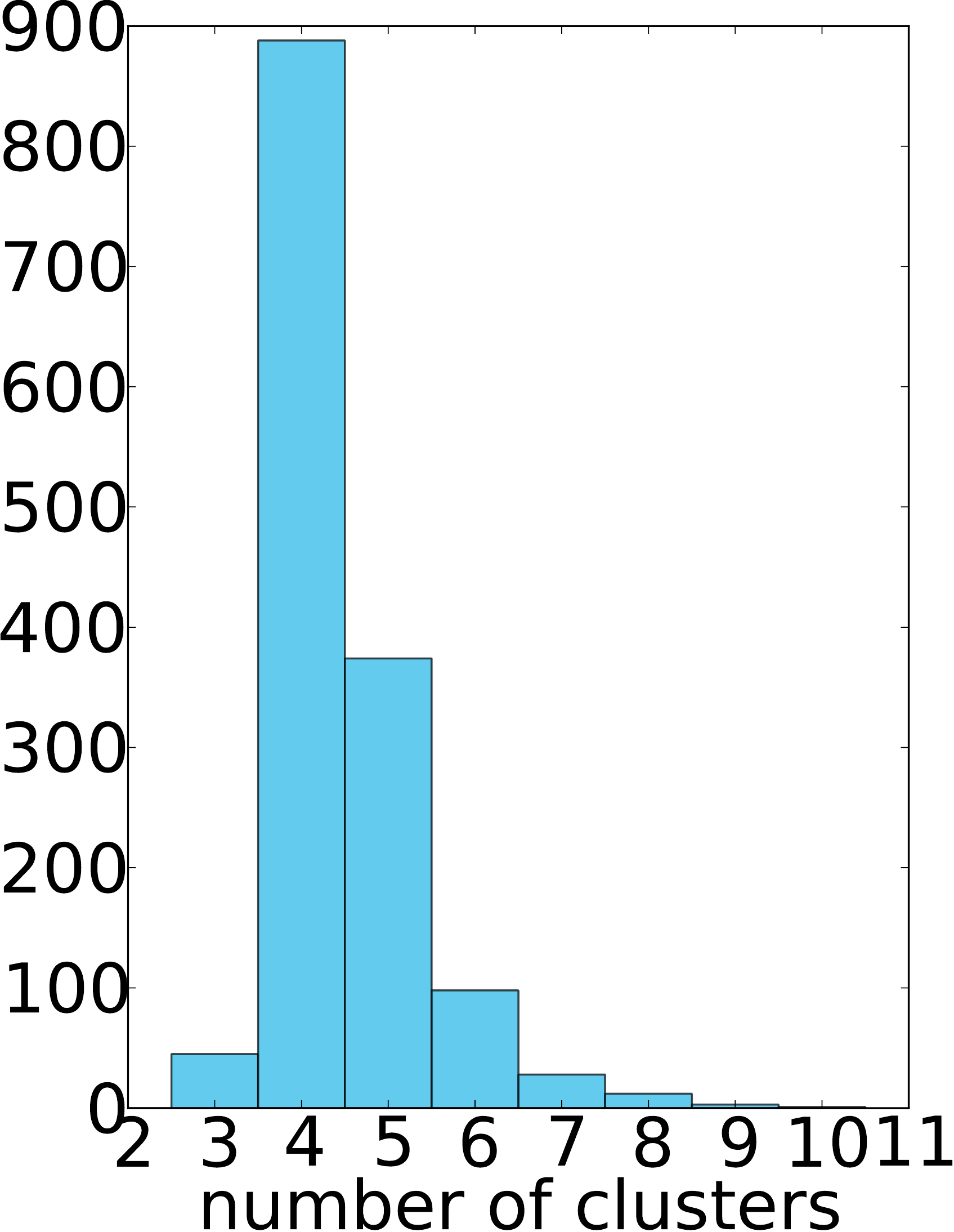}
  \end{subfigure}
  \begin{subfigure}[c]{0.30\textwidth}
    \includegraphics[width=\textwidth]{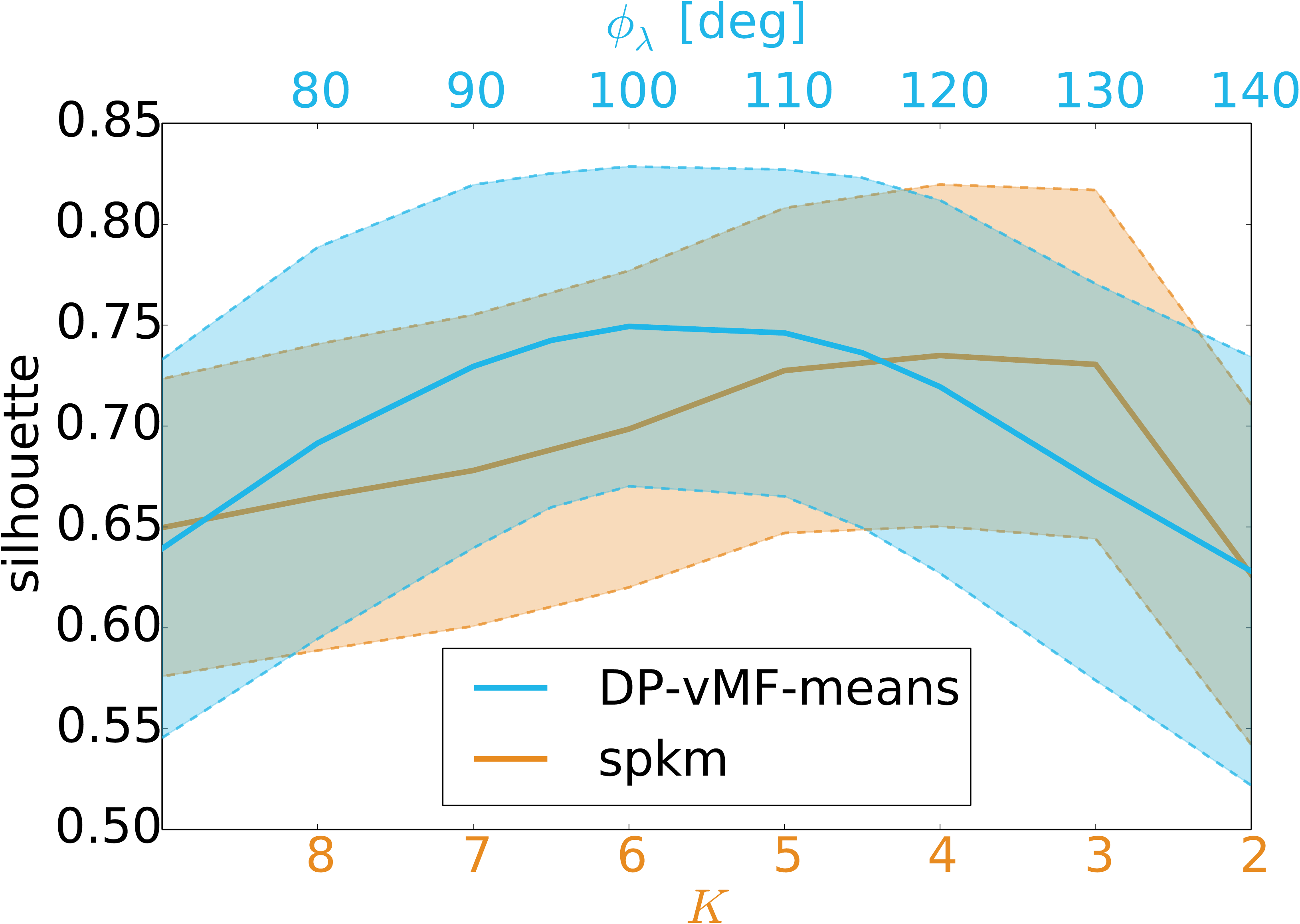}
  \end{subfigure}
  \caption{Histogram over the number of clusters found by
    DP-vMF-means (left) and silhouette values for DP-vMF-means and spkm
    (right) across the whole NYU dataset~\cite{Silberman:ECCV12}.\label{fig:nyuStats}}
\end{figure}

\section{Results}\label{sec:results}
\subsection{Evaluation of the DP-vMF-means Algorithm}

\begin{figure*}
  \centering
  \begin{tikzpicture}

    \draw (-7.0,1.1) node {
    \includegraphics[width=0.1338\linewidth]{./figures/office_kitchen_0001_409_rgb.png}
};
    \draw (-4.6,1.1) node {
    \includegraphics[width=0.1338\linewidth]{./figures/bathroom_0015_664_rgb.png}
};
    \draw (-2.2,1.1) node {
    \includegraphics[width=0.1338\linewidth]{./figures/living_room_0064_1314_rgb.png}
};
    \draw (0.2,1.1) node {
    \includegraphics[width=0.1338\linewidth]{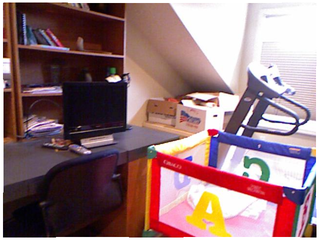}
};
    \draw (2.6,1.1) node {
    \includegraphics[width=0.1338\linewidth]{./figures/kitchen_0037_831_rgb.png}
};
    \draw (5.0,1.1) node {
    \includegraphics[width=0.1338\linewidth]{./figuresNYU/highK/home_office_0012_395_rgb.png}
};
    \draw (7.4,1.1) node {
    \includegraphics[width=0.1338\linewidth]{./figuresNYU/highK/bathroom_0028_691_rgb.png}
};

    \draw (-7.0,-0.7) node {
    \includegraphics[width=0.1338\linewidth]{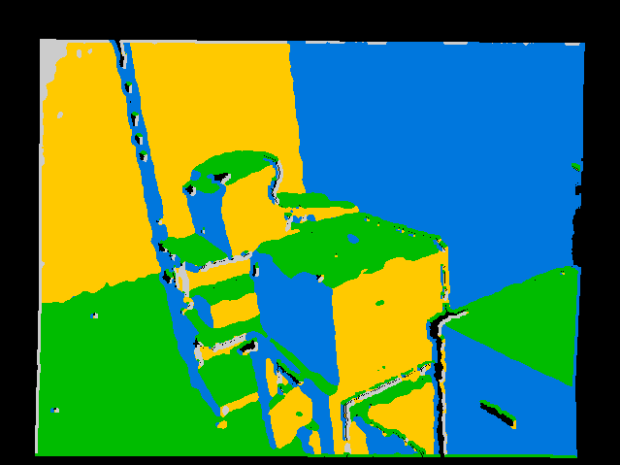}
};
    \draw (-4.6,-0.7) node {
    \includegraphics[width=0.1338\linewidth]{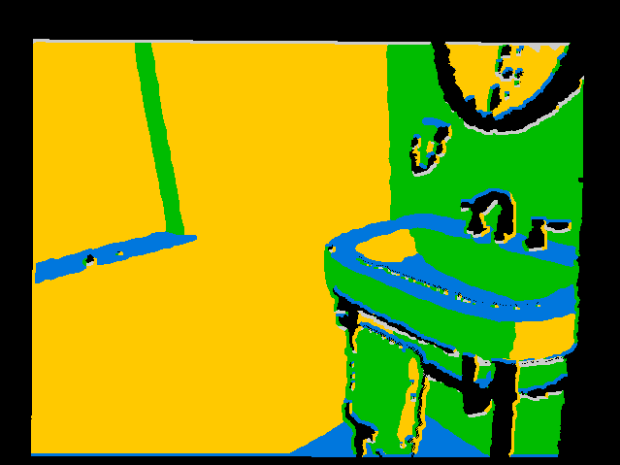}
};
    \draw (-2.2,-0.7) node {
    \includegraphics[width=0.1338\linewidth]{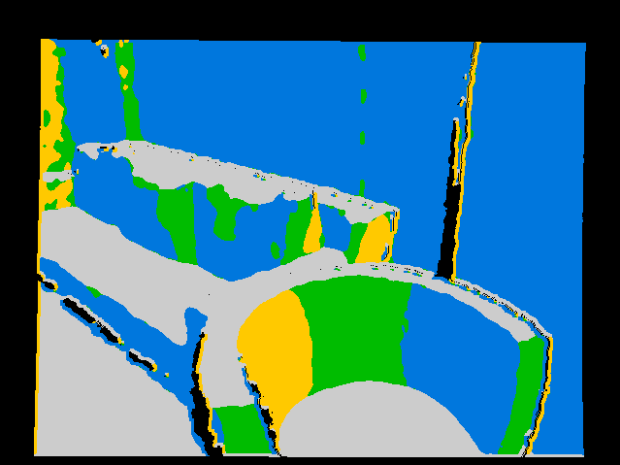}
};
    \draw (0.2,-0.7) node {
    \includegraphics[width=0.1338\linewidth]{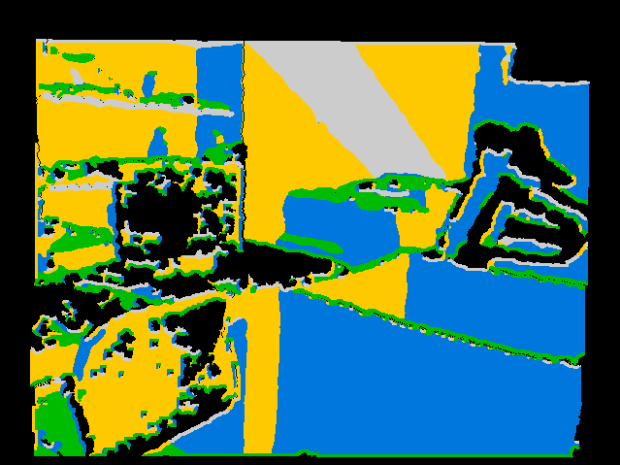}
};
    \draw (2.6,-0.7) node {
    \includegraphics[width=0.1338\linewidth]{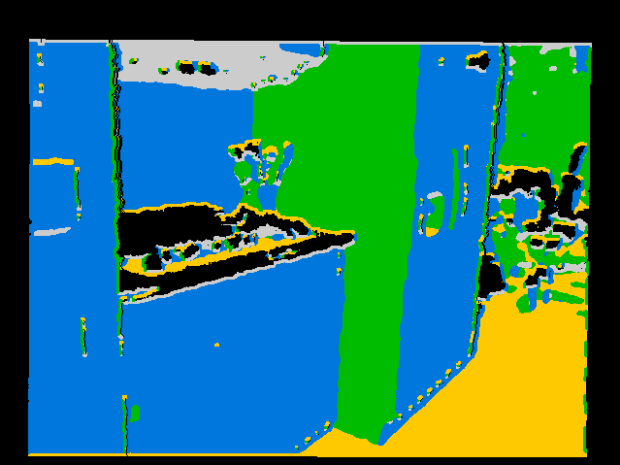}
};
    \draw (5.0,-0.7) node {
    \includegraphics[width=0.1338\linewidth]{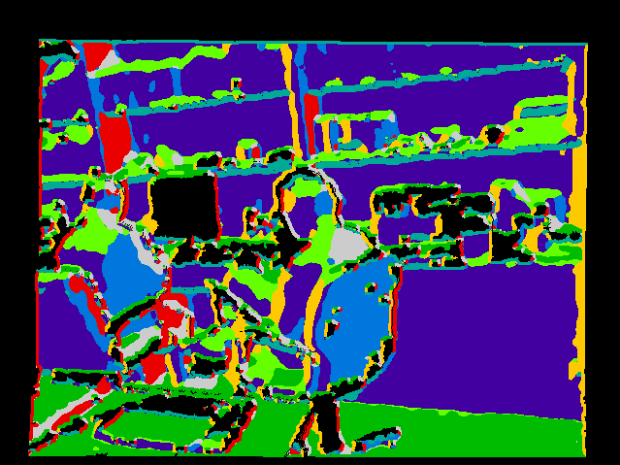}
};
    \draw (7.4,-0.7) node {
    \includegraphics[width=0.1338\linewidth]{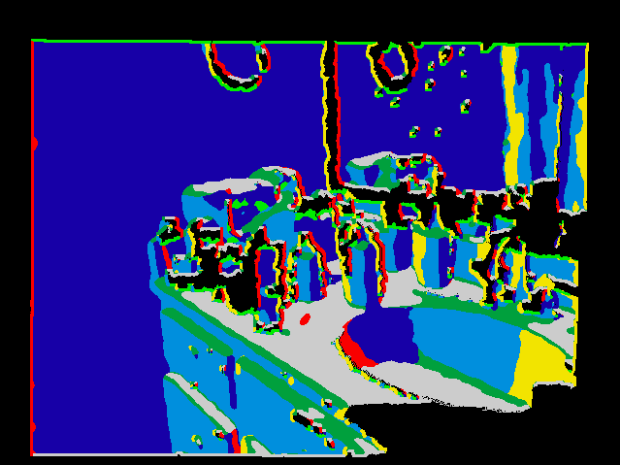}
};

    \draw (-7.0,-2.5) node {
    \includegraphics[width=0.1338\linewidth]{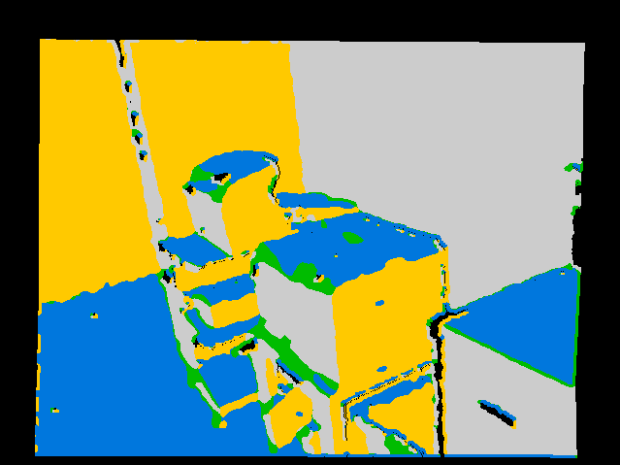}
};
    \draw (-4.6,-2.5) node {
    \includegraphics[width=0.1338\linewidth]{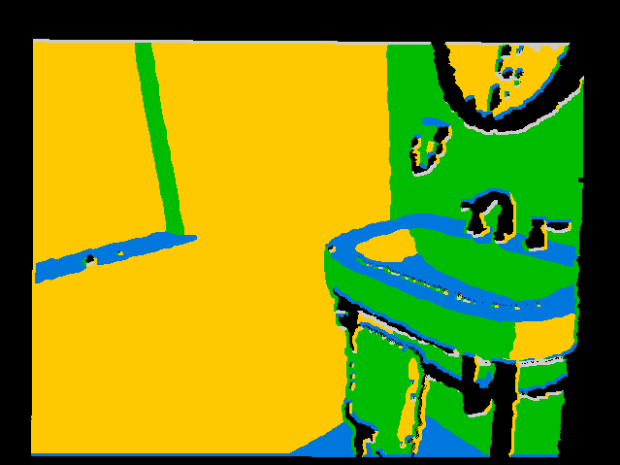}
};
    \draw (-2.2,-2.5) node {
    \includegraphics[width=0.1338\linewidth]{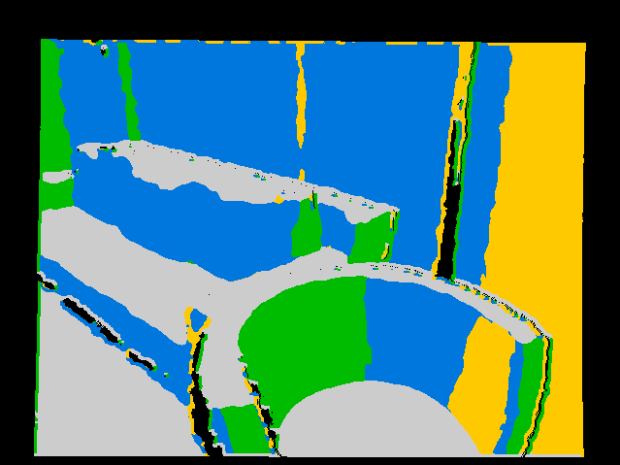}
};
    \draw (0.2,-2.5) node {
    \includegraphics[width=0.1338\linewidth]{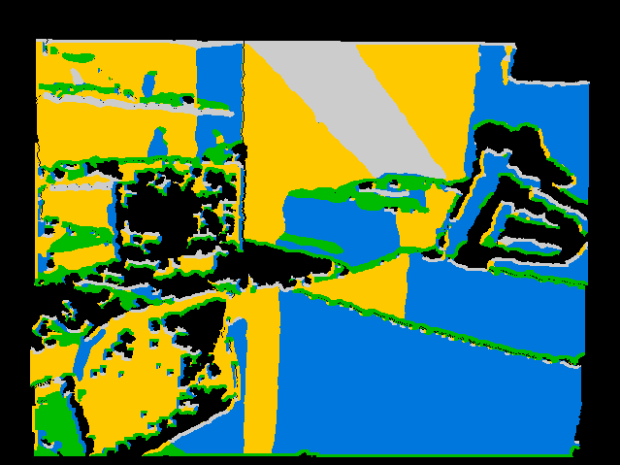}
};
    \draw (2.6,-2.5) node {
    \includegraphics[width=0.1338\linewidth]{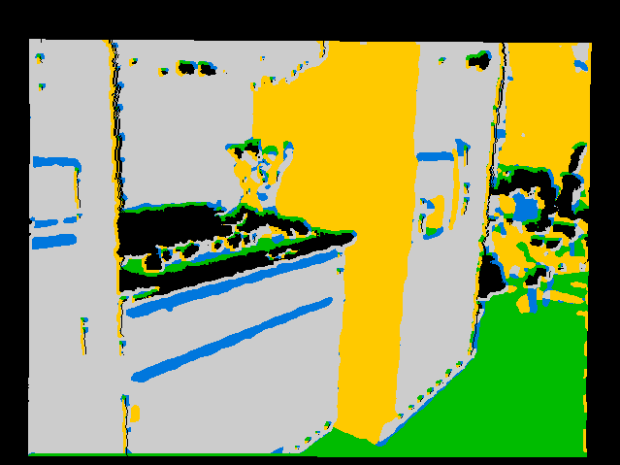}
};
    \draw (5.0,-2.5) node {
    \includegraphics[width=0.1338\linewidth]{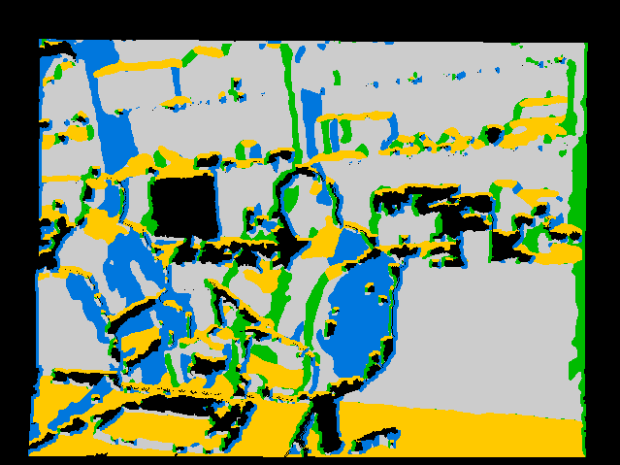}
};
    \draw (7.4,-2.5) node {
      \includegraphics[width=0.1338\linewidth]{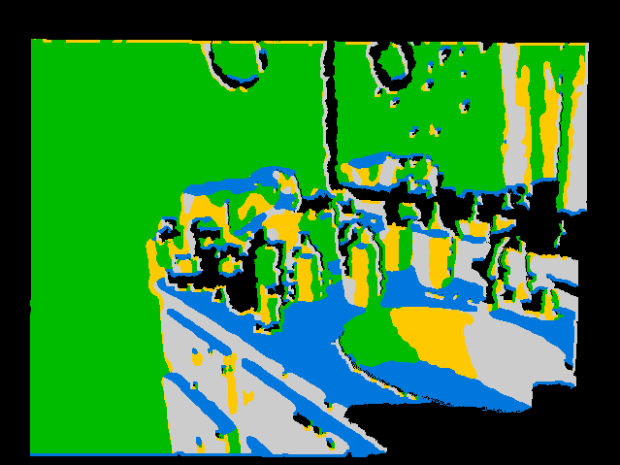}
};

    \draw (-7.0,-4.3) node {
    \includegraphics[width=0.1338\linewidth]{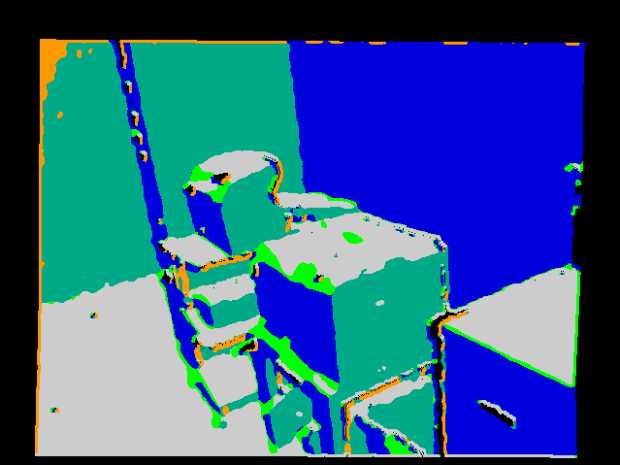}
};
    \draw (-4.6,-4.3) node {
    \includegraphics[width=0.1338\linewidth]{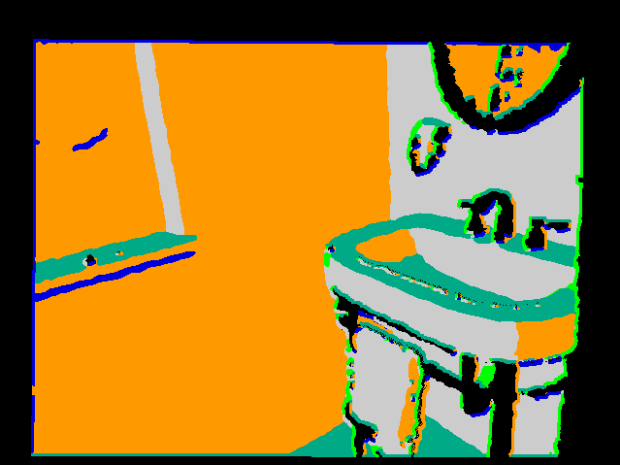}
};
    \draw (-2.2,-4.3) node {
    \includegraphics[width=0.1338\linewidth]{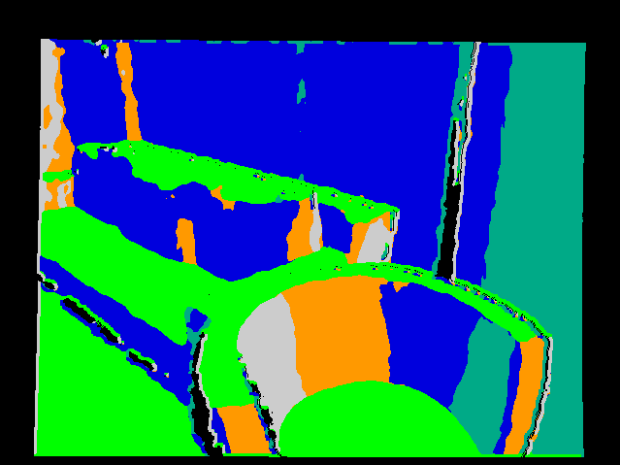}
};
    \draw (0.2,-4.3) node {
    \includegraphics[width=0.1338\linewidth]{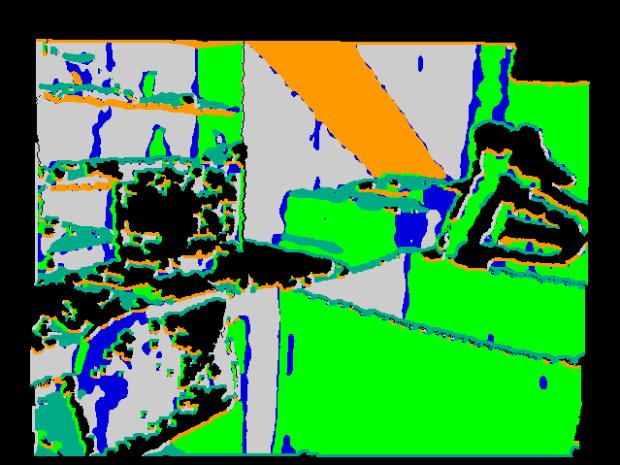}
};
    \draw (2.6,-4.3) node {
    \includegraphics[width=0.1338\linewidth]{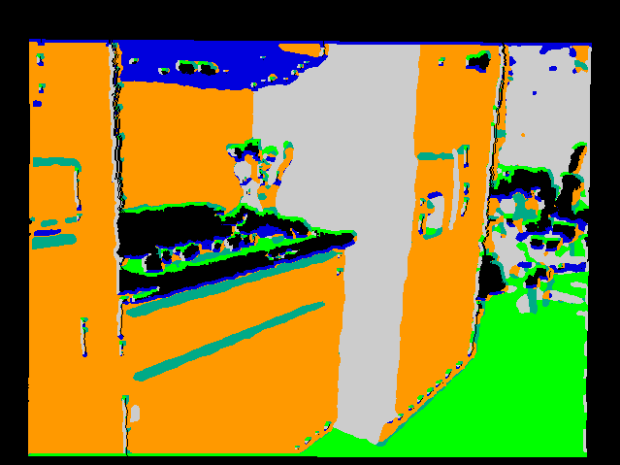}
};
    \draw (5.0,-4.3) node {
    \includegraphics[width=0.1338\linewidth]{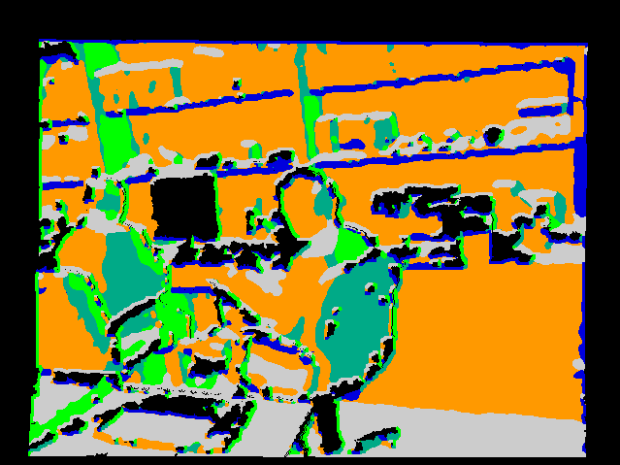}
};
    \draw (7.4,-4.3) node {
    \includegraphics[width=0.1338\linewidth]{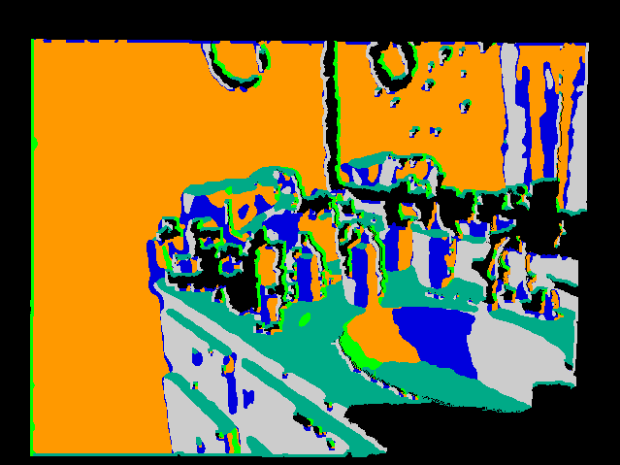}
};

\fill[gray] (-8.6,{1.1+0.86}) rectangle (-8.2,{1.1-0.86});
   \node[rotate=90] at (-8.4,1.1) {\scriptsize{RGB}};

   \fill[gray] (-8.6,{-0.7+0.89}) rectangle (-8.2,{-0.7-0.89});
   \node[rotate=90] at (-8.4,-0.7) {\scriptsize{DP-vMF-means}};

   \fill[gray] (-8.6,-1.62) rectangle (-8.2,-5.18);
   \node[rotate=90] at (-8.4,-4.3) {\scriptsize{spkm $K=5$}}; 
   \node[rotate=90] at (-8.4,-2.5) {\scriptsize{$K=4$}};

  \end{tikzpicture}
  \caption{Directional segmentation of scenes from the NYU v2 RGB-D
    dataset~\cite{Silberman:ECCV12} as implied by surface normal
    clusters. The complexity of the scenes increases from left to right
    as can be seen from the RGB images.  The second
    row shows the clustering inferred by DP-vMF-Means 
    while the third and fourth show the
    spherical \emph{k}-means results.
    Black denotes missing data due to sensor limitations.
  Note that \DPvMFmeans{} adapts the number of clusters to the
complexity of the scene.  \label{fig:nyuQualitative}}
\end{figure*}
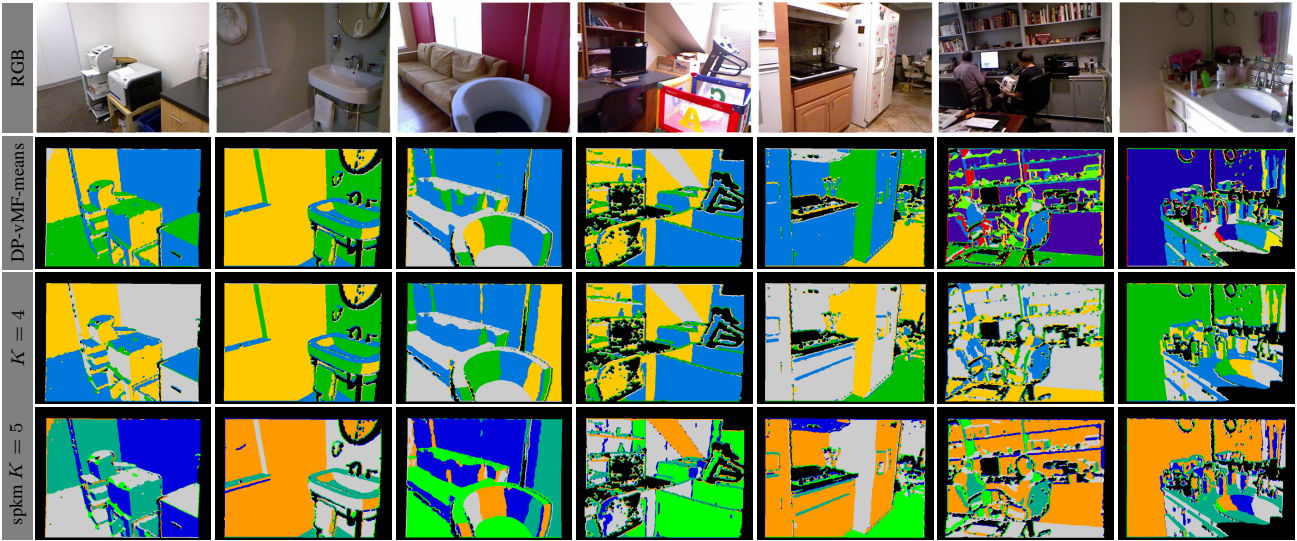

\begin{wrapfigure}{r}{0.20\columnwidth}
  \vspace{-.4cm}
  \hspace{-.55cm}\includegraphics[width=0.25\columnwidth]{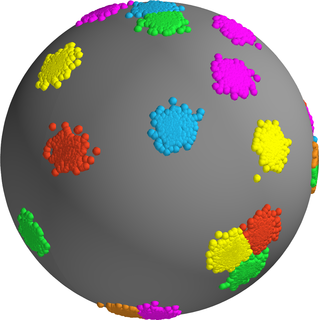}
    \caption*{\hspace{-.6cm}Synthetic Data}
  \vspace{-.4cm}
\end{wrapfigure}
\textbf{Synthetic Data:}
First, we evaluate the behavior of the \DPvMFmeans{} algorithm in
comparison to its parametric cousin, the \spkm{} algorithm, on
synthetic 3D spherical data sampled from $K_\text{T} = 30$ true vMF
distributions. All evaluation results are shown as the mean and
standard deviation over 50 runs.  The left plot of
Fig.~\ref{fig:synthEval} depicts the inferred number of clusters $K$ on
the horizontal axis as a function of the respective parameters of the
two algorithms: the number of clusters $K$ for \spkm{} and the
parameter $\phi_\lambda$ for \DPvMFmeans{} (recall that $\phi_\lambda =
\cos^{-1}(\lambda+1)$ as defined in Sec.~\ref{sec:dpvmf}).  This figure
demonstrates the ability of the DP-vMF-means algorithm to discover the
correct number of clusters $K_\text{T}$, and the relative insensitivity
of the discovered number of clusters with respect to its parameter
$\phi_\lambda$.

The middle and right hand plots show two measures for clustering
quality.  The normalized Mutual Information
(NMI)~\cite{strehl2003cluster}, depicted in the middle, is computed
using the true labels.  \DPvMFmeans{} achieves an almost perfect NMI of
$0.99$, while \spkm{} only reaches $0.94$ NMI even with $K=K_\text{T}$.
The slightly superior performance of DP-vMF-means stems from its
enhanced ability to avoid local optima due to the way labels are
initialized: while \spkm{} is forced to initialize $K$ cluster
parameters, \DPvMFmeans{} starts with an empty set and adds clusters on
the fly as more data are labelled.  
The NMI results are corroborated by the silhouette
score~\cite{rousseeuw1987silhouettes}, shown to the right in
Fig.~\ref{fig:synthEval}.  
The silhouette
score is an internal measure for clustering quality that can be
computed without knowledge of the true clustering, and is used
to tune parametric clustering algorithms.
With a
maximum of $0.92$ \DPvMFmeans{} reaches a close to perfect silhouette
score, indicating well-separated, concentrated clusters. Again, \spkm{}
does not reach the same clustering performance even for $K=K_\text{T}$
for the same aforementioned reasons.

\begin{figure*}
  \centering
  \begin{tikzpicture}[scale=\textwidth/17.8cm]
    \def\imgW{2.30cm}
    \def\barW{ {2.30cm*7+0.025cm*14} }
    \draw (-7.0,0.88) node {\includegraphics[width=\imgW]{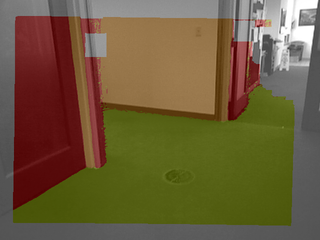}};
    \draw (-4.6,0.88) node {\includegraphics[width=\imgW]{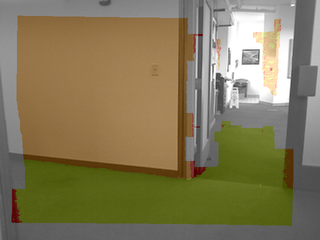}};
    \draw (-2.2,0.88) node {\includegraphics[width=\imgW]{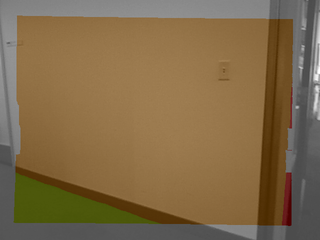}};
    \draw (0.2,0.88) node {\includegraphics[width=\imgW]{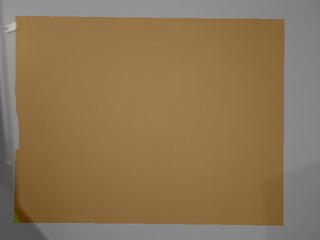}};
    \draw (2.6,0.88) node {\includegraphics[width=\imgW] {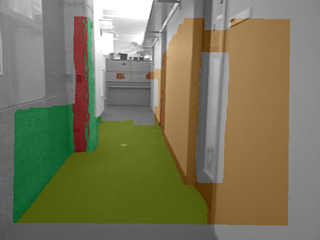}};
    \draw (5.0,0.88) node {\includegraphics[width=\imgW] {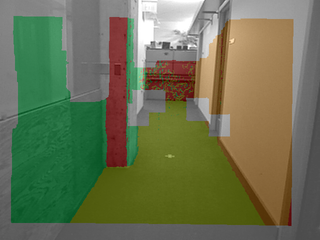}};
    \draw (7.4,0.88) node {\includegraphics[width=\imgW] {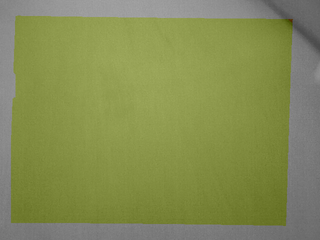}};

    \draw (0.2,-0.42) node {\includegraphics[width=\barW,clip,trim=0 0 25 5]{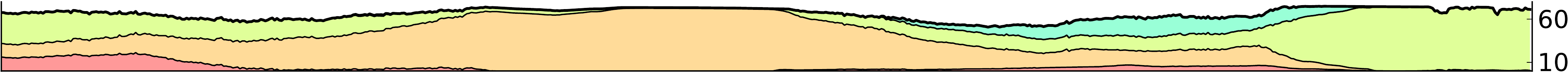}};
    \draw (0.2,-1.2) node {\includegraphics[width=\barW,clip,trim=0 0 25 5]{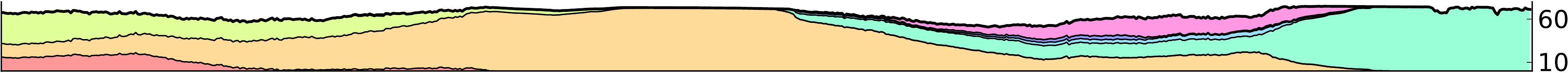}};
    \fill[gray] ({8.8+0.2},{-0.812+.748}) rectangle ({8.8-0.2},{-0.812-.748});
    \node[rotate=90] at (8.8,-0.812) {\scriptsize{cluster shares}};

    \fill[gray] ({8.8+0.2},{0.88+0.88}) rectangle ({8.8-0.2},{0.88-0.88});
    \node[rotate=90] at (8.8,0.88) {\scriptsize{segmentation}};

    \fill[gray] ({8.8+0.2},{-2.5+0.88}) rectangle ({8.8-0.2},{-2.5-0.88});
    \node[rotate=90] at (8.8,-2.5) {\scriptsize{segmentation}};

    \fill[gray] ({8.8+0.2},{-4.3+0.88}) rectangle ({8.8-0.2},{-4.3-0.88});
    \node[rotate=90] at (8.8,-4.3) {\scriptsize{segmentation}};

    \draw (-7.0,-2.5) node {\includegraphics[width=\imgW]{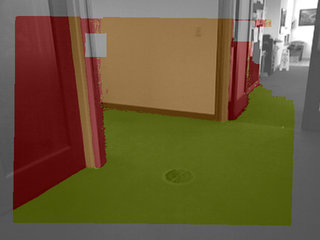}};
    \draw (-4.6,-2.5) node {\includegraphics[width=\imgW]{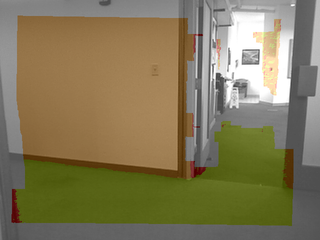}};
    \draw (-2.2,-2.5) node {\includegraphics[width=\imgW]{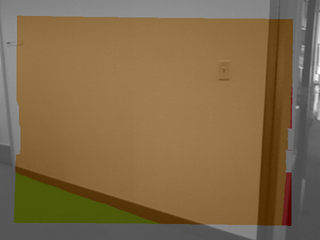}};
    \draw (0.2,-2.5) node {\includegraphics[width=\imgW]{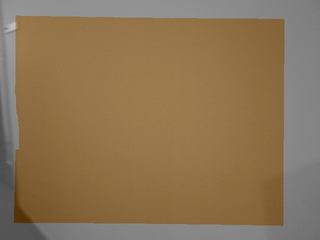}};
    \draw (2.6,-2.5) node {\includegraphics[width=\imgW] {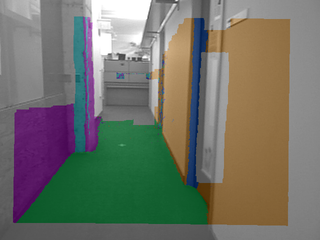}};
    \draw (5.0,-2.5) node {\includegraphics[width=\imgW] {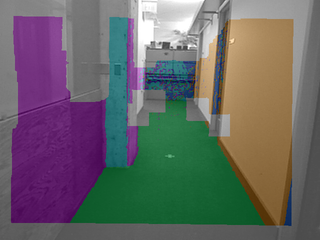}};
    \draw (7.4,-2.5) node {\includegraphics[width=\imgW] {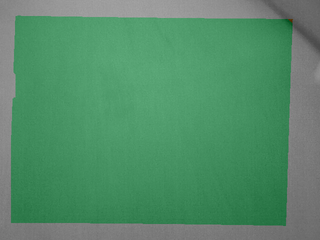}};

    \draw (-7.0,-4.3) node {\includegraphics[width=\imgW]{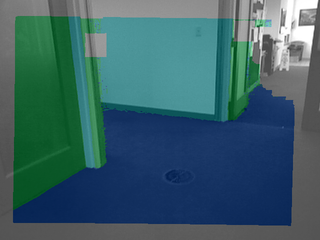}};
    \draw (-4.6,-4.3) node {\includegraphics[width=\imgW]{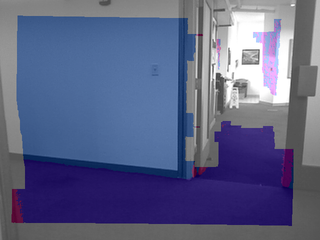}};
    \draw (-2.2,-4.3) node {\includegraphics[width=\imgW]{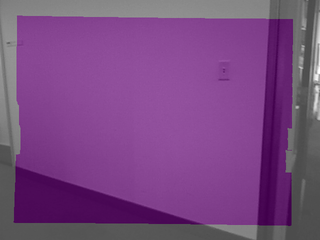}};
    \draw (0.2,-4.3) node {\includegraphics[width=\imgW]{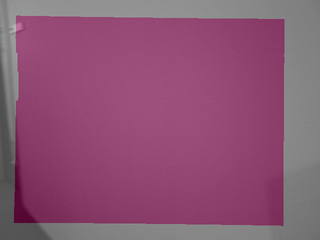}};
    \draw (2.6,-4.3) node {\includegraphics[width=\imgW] {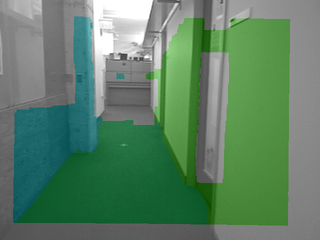}};
    \draw (5.0,-4.3) node {\includegraphics[width=\imgW] {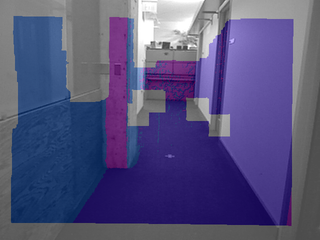}};
    \draw (7.4,-4.3) node {\includegraphics[width=\imgW] {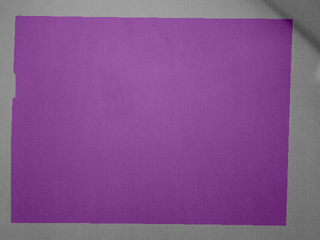}};

   \fill[gray] (-8.6,1.76) rectangle (-8.2,-0.78);
   \node[rotate=90] at (-8.4,0.49) {\scriptsize{DDP-vMF-means}};

   \fill[gray] (-8.6,-0.84) rectangle (-8.2,-3.38);
   \node[rotate=90] at (-8.4,-2.1) {\scriptsize{sDP-vMF-means}};

   \fill[gray] (-8.6,-3.42) rectangle (-8.2,-5.18);
   \node[rotate=90] at (-8.4,-4.3) {\scriptsize{DP-vMF-means}};

   \draw[->,thick,black] (-8.170,-5.30) to (8.57,-5.30);
   \foreach \x in {-7,-4.6,-2.2,0.2,2.6,5.0,7.4}
     \draw[thick,black] (\x,-5.30) -- (\x,-5.35); 
   \draw[anchor=north] (-7.0,-5.30) node {\scriptsize 30};
   \draw[anchor=north] (-4.6,-5.30) node {\scriptsize 120};
   \draw[anchor=north] (-2.2,-5.30) node {\scriptsize 210};
   \draw[anchor=north] ( 0.2,-5.30) node {\scriptsize 330};
   \draw[anchor=north] ( 2.6,-5.30) node {\scriptsize 420};
   \draw[anchor=north] ( 5.0,-5.30) node {\scriptsize 510};
   \draw[anchor=north] ( 7.4,-5.30) node {\scriptsize 600};

   \node[anchor=north] at (-0.9,-5.30) {\scriptsize{frame number}};

  \end{tikzpicture}
  \caption{Clustering of a surface normal stream recorded when walking
    a $90^\circ$ turn in an office environment. We depict key-frames
    color-coded with the implied surface-normal clustering for three
    clustering algorithms. The plots in the second and third
    row depict the percentage of normals associated to the respective
    cluster for DDP-vMF-means and sequential DP-vMF-means.
    Note that only the clustering obtained via the DDP-vMF-means
    algorithm is consistent across the whole run.
\label{fig:ddpvmf}}
\end{figure*}

For additional reference we ran two sampling-based inference
  algorithms for the DP-vMF-MM model. 
  The CRP-based inference of~\cite{bangert2010using} 
  was aborted after running for two days
  without convergence. 
  This inefficiency has been noted
  previously~\cite{Jain00asplit-merge}.
A more efficient alternative is the finite Dirichlet process (FSD)
approximation~\cite{ishwaran2002exact} to the DP-vMF-MM. The inference
can be parallelized and yielded results within minutes.
After $5000$ iterations the sampler converged to an incorrect number of
$23.0 \pm 1.67$ clusters on average over 10 sampler runs. These
convergence issues have also been noted before~\cite{chang13dpmm}. The
last sample of the different runs achieves an average NMI of $0.94 \pm
0.01$ and a silhouette score of $0.74\pm 0.04$. 

\textbf{NYU v2 depth dataset:} 
In this experiment, the \DPvMFmeans{} and \spkm{} algorithms were
compared on the NYU v2 RGB-D dataset~\cite{Silberman:ECCV12}.  Surface
normals were extracted from the depth
images~\cite{Holz11RoboCup} and preprocessed with total variation
smoothing~\cite{RosmanWTKB12}.
We quantify the clustering quality in terms of the average silhouette
score over the clusterings of the 1449 scenes of the NYU v2 depth
dataset.  Since we do not possess the true scene labeling, we use the
silhouette quality metric as a proxy for the NMI metric; this was
motivated by results of the synthetic experiment.

Across the whole NYU v2 dataset, the \DPvMFmeans{} algorithm achieves
the highest average silhouette score of $0.75$ for $\phi_\lambda^\star
= 100^\circ$ as depicted in Fig.~\ref{fig:nyuStats}.  The histogram
over the number of inferred clusters by \DPvMFmeans{} for
$\phi_\lambda^\star$ indicates the varying complexity of the scenes
ranging from three to eleven.  The clear peak at $K=4$ coincides with
the highest silhouette score for \spkm{} ($0.73$) and explains the only
slightly lower silhouette score of \spkm{}: most scenes in the dataset
exhibit four primary directions. 

Figure~\ref{fig:nyuQualitative} shows a qualitative comparison of the
scene segmentation implied by the clustering of surface normals. In
comparison to \spkm{}, the \DPvMFmeans{} clustering results show the
ability of the algorithm to adapt the number of clusters to the scene
at hand. If the right number of clusters is selected for the \spkm{}
clustering, the results have similar quality; however, the number of
clusters is generally not known a priori
and varies across scenes.  This demonstrates two major advantages of
\DPvMFmeans{} over \spkm{}: (1) \DPvMFmeans{} is less sensitive to the
parameter setting (see Fig,~\ref{fig:synthEval}, left) and (2) it is
easier to choose $\phi_\lambda$ than $K$ since it intuitively
corresponds to the maximum angular radius of a cluster, which can be
gauged from the type of data and its noise characteristics.  For this
experiment $\phi_\lambda=100^\circ$ is justified by the typical
Manhattan structure~\cite{coughlan1999manhattan} of the indoor environment plus $10^\circ$ to
account for sensor noise.

\subsection{Evaluation of the DDP-vMF-means Algorithm}
\paragraph{Real-time Directional Segmentation:}

In fields such as mobile robotics or augmented reality, it is uncommon
to observe just a single RGB-D frame of a scene; more typically, the
sensor will observe a temporal sequence of frames. The following
experiment demonstrates the temporally consistent clustering capability
of the DDP-vMF-means algorithm on surface normals extracted from a sequence
of depth images recorded in an indoor environment. Each frame is 
preprocessed in $11$ms using edge-preserving smoothing
with a hybrid CPU-GPU guided filter~\cite{he2010guided}.

We compare against the ad-hoc approaches of clustering on a
frame-by-frame basis using DP-vMF-means, both with and without
initializing the algorithm from the previous frame's clusters. The
former is referred to as sequential DP-vMF-means (sDP-vMF-means).
sDP-vMF-means achieves a greedy frame-to-frame label consistency,
but, unlike DDP-vMF-means, it cannot reinstantiate previous clusters after
multiframe lapses.
Motivated by the DP-vMF-means evaluation, all algorithms were run
with $\phi_\lambda=100^\circ$. For DDP-vMF-means $\beta = 10^5$ and $Q
=\frac{\lambda}{400}$.

The differences in labeling consistency can be observed in rows
two and three of Fig.~\ref{fig:ddpvmf}, which shows the percentage of
normals associated with a specific cluster. While DDP-vMF-means is
temporally consistent and reinstantiates the lime-green and red
clusters, observed in the first half of the run, DP-vMF-means
erroneously creates new clusters.  We do not depict the percentages of
surface normals associated with the clusters for the batch DP-vMF-means
algorithm, since the there is no label consistency between time-steps
as can be observed in the last row of Fig.~\ref{fig:ddpvmf}.

The average run-time per frame was $28.4$~ms for batch DP-vMF-means,
$12.8$~ms for sDP-vMF-means, $20.4$~ms for DDP-vMF-means, and $13.6$~ms for spkm
with $K=5$.
The increased running time of batch DP-vMF-means is a result of
clustering each batch of surface normals in isolation; OIR label
assignment needs several restarts to
assign labels to all surface normals.  By initializing the clusters from a
previous frame, sDP-vMF-means only incurs labeling restarts if a
new cluster is observed, and hence has significantly lower run time.
DDP-vMF-means is slightly slower than sDP-vMF-means since it is
keeping track of both observed and unobserved clusters.

\section{Conclusion}\label{sec:conclusion}
Taking the small-variance asymptotic limit of the Bayesian
nonparametric DP-vMF and DDP-vMF mixture models, we have derived two
novel spherical kmeans-like algorithms for efficient batch and
streaming clustering on the unit hypersphere.  The performance and
flexibility of DP-vMF-means was demonstrated on both synthetic data and
the NYU~v2 RGB-D dataset. For DDP-vMF-means, Optimistic Iterated
Restarts (OIR) parallelized label assignments, enable real-time
temorally consistent clustering of batches of 300k surface normals collected at 30~Hz from a
RGB-D camera.

We envision a large number of potential applications for the presented
algorithms in computer vision and in other realms where directional
data is encountered. Implementations are 
available at {\normalsize\url{http://people.csail.mit.edu/jstraub/}}.

\subsubsection*{Acknowledgements}
This work was partially supported by ONR MURI N00014-11-1-0688 and
ARO MURI W911NF-11-1-0391.
%We would additionally like to thank the free coffee, our whiteboard and the openness of LIDS and CSAIL.

%\subsubsection*{References}
%
%References follow the acknowledgements.  Use an unnumbered third level
%heading for the references section.  Any choice of citation style is
%acceptable as long as you are consistent.  Please use the same font
%size for references as for the body of the paper---remember that
%references do not count against your page length total.
%
\end{document}